\documentclass[11pt]{article}
\usepackage[margin=1in]{geometry}

\usepackage{xcolor}
\usepackage[table]{xcolor}
\usepackage{xspace}

\usepackage{bm}        
\usepackage{mathtools} 
\usepackage{amsmath}
\usepackage{amssymb}
\usepackage{amsfonts}
\usepackage{amsthm}

\usepackage{booktabs}   
\usepackage{graphicx}   
\usepackage{makecell}   

\newcommand{\E}{\mathbb{E}}

\newcommand{\OLO}{\texttt{OLO}}
\newcommand{\OSM}{\texttt{OSM}}

\usepackage[margin=1in]{geometry}
\usepackage{amsmath,amssymb,amsthm,mathtools}
\usepackage{enumitem}
\usepackage{algorithm}
\usepackage{algpseudocode}
\usepackage{hyperref}
\usepackage{natbib}
\usepackage{xcolor}
\usepackage{comment}

\newtheorem{lemma}{Lemma}
\newtheorem{theorem}{Theorem}
\newtheorem{corollary}{Corollary}

\newtheorem*{restatedtheorem}{Theorem}
\newtheorem*{restatedlemma}{Lemma}
\newtheorem*{restatedcorollary}{Corollary}

\newcommand{\calA}{\mathcal{A}}

\newcommand{\calO}{\mathcal{O}}

\newcommand{\calR}{\mathcal{R}}

\newcommand{\bone}{\boldsymbol{1}}

\newcommand{\bg}{\bm{g}}

\newcommand{\bp}{\bm{p}}
\newcommand{\bq}{\bm{q}}
\newcommand{\br}{\bm{r}}
\newcommand{\bs}{\bm{s}}

\newcommand{\bu}{\bm{u}}
\newcommand{\bv}{\bm{v}}
\newcommand{\bw}{\bm{w}}

\newcommand{\by}{\bm{y}}
\newcommand{\bz}{\bm{z}}

\newcommand{\bH}{\bm{H}}

\def\<{\langle}
\def\>{\rangle}

\newcommand{\spara}[1]{\smallbreak\noindent{\bf{#1}}}

\newcommand{\squishlist}{\begin{list}{$\bullet$}
  { \setlength{\itemsep}{0pt}
     \setlength{\parsep}{3pt}
     \setlength{\topsep}{3pt}
     \setlength{\partopsep}{0pt}
     \setlength{\leftmargin}{1.5em}
     \setlength{\labelwidth}{1em}
     \setlength{\labelsep}{0.5em} } }
\newcommand{\squishend}{
  \end{list}  }

\DeclareMathOperator{\Reg}{\calR}
\newcommand{\aReg}{\Reg_T^{1-1/e}}

\newcommand{\Random}{\texttt{Random}\xspace}
\newcommand{\MSEThree}{\texttt{MSE3}\xspace}
\newcommand{\OpaqueGreedy}{\texttt{OpaqueGreedy}\xspace}
\newcommand{\BanditIG}{\texttt{Bandit-IG}\xspace}
\newcommand{\alg}{\texttt{Sub-EW}}
\newcommand{\algfull}{\texttt{Submodular-Exponential-Weights}}

\newcommand{\BanditMLSM}{\texttt{BanditMLSM}\xspace}
\newcommand{\PoissonBE}{\texttt{Poisson-BE}\xspace}

\newcommand{\Influence}{\texttt{Influence}\xspace}
\newcommand{\TeamFormation}{\texttt{TeamFormation}\xspace}
\newcommand{\Coverage}{\texttt{Coverage}\xspace}

\newcommand{\problem}{\textit{Bandit-OSM}}

\hypersetup{hidelinks}

\makeatletter
\@ifpackageloaded{natbib}{}{\usepackage[round]{natbib}}
\makeatother

\title{Efficient Online Submodular Maximization
       with Bandit Feedback}

\author{
  Iasonas Nikolaou\\
  Boston University\\
  \texttt{nikolaou@bu.edu}
  \and
  Evimaria Terzi\\
  Boston University\\
  \texttt{evimaria@bu.edu}
}
\date{}

\begin{document}
\maketitle

\begin{abstract}
We study online monotone submodular maximization under a cardinality constraint with bandit feedback. On each round, the learner selects at most $k$ elements from a ground set of size $n$ and observes only the value of the selected set. 
We develop an efficient algorithm achieving $\calO\!\left(k\sqrt{nT\log(n/k)}\right)$ expected $(1-1/e)$-regret over $T$ rounds. To our knowledge, this is the first polynomial-time algorithm with $\sqrt{T}$ dependence on the horizon for arbitrary monotone submodular functions in the bandit setting. 
Our approach uses randomized sampling and a carefully constructed potential to reduce online submodular maximization to online linear optimization over the probability simplex.
\end{abstract}

\section{Introduction}
\label{sec:introduction}
Submodular optimization is a central theme in machine learning.  The key reason for that  is that
submodularity captures diminishing returns and arises in numerous applications including sensor placement~\citep{krause2007near}, team formation~\citep{bhowmik2014submodularity}, caching networks~\citep{li2021online}, influence maximization~\citep{kempe2003maximizing}, recommender systems~\citep{nikolaou2026online}, and multi-robot planning~\citep{xu2023online}.
Motivated by these applications, a growing literature studies \emph{Online Submodular Maximization (\OSM)} under various settings~\citep{streeter2009online,zhang2019online,zhang2022stochastic, salem2024online}.

In {\OSM}, a learner, at each round \(t\), selects a set \(S_t\subseteq V\) with \(|S_t|\leq k\) and observes 
a monotone submodular reward $f_t$. The goal is to minimize the expected \emph{\((1-1/e)\)-regret}, i.e.,
\[
\aReg \coloneqq
\left(1-\frac{1}{e}\right)
\max_{\substack{S\subseteq V\\ |S|\leq k}}
\sum_{t=1}^{T} f_t(S)
-
\mathbb{E}\!\left[\sum_{t=1}^{T} f_t(S_t)\right].
\]
This quantity compares the learner's cumulative reward with that of the best fixed set in hindsight. The $(1-1/e)$ factor is necessary for polynomial-time algorithms due to 
the hardness results associated with the offline problem~\citep{feige1998threshold}.

In this paper, we focus on \emph{Bandit-OSM}, where at every round $t$ the learner only observes the value of the function on the selected set, i.e., $f_t(S_t)$ 
(as opposed to receiving the whole $f_t$). 
The best existing results for {\problem} achieve
$\aReg = \widetilde{\calO}(T^{2/3})$ both for cardinality~\citep{wan2023bandit} and for general matroid constraints~\citep{wan2026bandit}. 

This raises the question:
\begin{quote}
\emph{Is it possible to obtain a computationally efficient algorithm
for {\problem} with $\calO(\sqrt{T})$ expected $(1-1/e)$-regret?}
\end{quote}
We answer this question affirmatively: to our knowledge, we develop the first polynomial-time algorithm that achieves $\calO(\sqrt{T})$ expected $(1-1/e)$-regret
for {\problem} under cardinality constraints with arbitrary monotone
submodular reward functions.

Our algorithm, {\algfull} ({\alg}), is a bandit exponential-weights method based on EXP3~\citep{auer2002nonstochastic}. 
{\alg} is conceptually simple:
it maintains a distribution over the elements of \(V\). At each iteration, it randomly chooses how many elements to sample, draws them independently, and plays their union. Randomizing the number of draws makes it possible to estimate, from the observed reward, the gradient of a carefully-constructed potential.
The resulting gradient estimate guides the exponential-weights update of the distribution over \(V\).

\section{Related work}
\label{sec:related-work}

In this section, we give an overview of the existing results on Online Submodular Maximization (OSM) in the \emph{bandit} setting (i.e., {\problem}).

{\problem} is considerably more
challenging than its full-information counterpart: the learner
observes only the value of the selected set and cannot directly
evaluate marginal gains. Early work by~\citet{streeter2008online}
introduced the meta-action approach for online monotone
submodular maximization under a cardinality constraint. Their
bandit construction simulates the steps of the offline greedy
algorithm using separate online learners and achieves $\aReg = \widetilde{\calO}(T^{2/3})$.
\citet{streeter2009online} extended this approach to online
assignment problems, a special case of partition matroid
constraints, obtaining regret $\aReg$ of order $\widetilde{\calO}(T^{4/5})$.

\citet{niazadeh2023online} developed a general offline-to-online
reduction based on Blackwell approachability. Their framework
transforms robust greedy approximation algorithms into online
algorithms, yielding the bound $\aReg=\calO(T^{2/3})$ for Bandit-OSM.
In a different direction,~\citet{zhang2019online}
introduced Bandit--Frank--Wolfe for online continuous
DR-submodular maximization under bandit feedback, attaining
$\aReg=\calO(T^{8/9})$.
Their discrete counterpart, Responsive--Frank--Wolfe, uses
responsive-bandit feedback, requiring evaluations at infeasible
sets that yield zero reward.

\citet{wan2023bandit} reduced {\problem} to bandit multilinear
DR-submodular maximization, achieving
$\aReg=\widetilde{\calO}(T^{2/3})$ under partition matroid constraints.
More recently,~\citet{wan2026bandit} obtained the expected regret
bound $\aReg=\widetilde{\calO}(T^{2/3})$ for {\problem} under
general matroid constraints.

As summarized in Table~\ref{tab:bandit-related-work}, none of the above works
achieves the regret bound $\aReg=\calO(\sqrt{T})$
under cardinality constraints for {\problem}, which is what we establish here.

To the best of our knowledge, the only other polynomial-time result achieving
$\aReg=\widetilde{\calO}(\sqrt{T})$ under bandit feedback and
cardinality constraints is due to \citet{pasteris2023sum}.
Their analysis applies to monotone and submodular objectives that additionally satisfy pseudo-concavity, including
\emph{sum-max functions}.
This class captures facility-location and $k$-median rewards,
coverage and hitting-set objectives, and best-of-$k$ bandits,
but is a proper subclass of monotone submodular functions.
In contrast, our guarantee holds for arbitrary monotone
submodular rewards.
At each round, their algorithm samples $k$ elements with
replacement and plays their union.
The observed reward is used to construct a gradient estimate
for an exponential-weights update.
The pseudo-concavity property yields a concave
expected-reward potential.
In contrast, we construct a weighted potential that combines expected
rewards from different numbers of independent draws with replacement,
ranging from one to $k$.
This potential allows us to handle arbitrary monotone submodular
functions without assuming pseudo-concavity.
By randomizing the number of draws, we obtain an unbiased estimate
of the potential's gradient from a single observed reward while
retaining the $(1-1/e)$ approximation factor.

\paragraph{Stochastic submodular bandits.}
A related literature considers a variant of our setting where the learner receives bandit feedback, but the reward functions are chosen \emph{stochastically} (i.e.,  not adversarially).
\citet{nie2022explore} proposed an explore-then-commit
greedy algorithm for 
this setting and 
stochastic monotone submodular
rewards and cardinality constraints and established
$\aReg=\widetilde{\calO}(T^{2/3})$.
\citet{tajdini2024nearly} subsequently gave nearly
minimax-optimal upper and lower bounds for a stronger regret
notion based on robust-greedy maximization.

\begin{table*}[t]
\centering
\scriptsize
\renewcommand{\arraystretch}{1.25}
\setlength{\tabcolsep}{5pt}
\resizebox{\textwidth}{!}{
\begin{tabular}{@{}lcccc@{}}
\toprule
\textbf{Paper}
& \textbf{Setting}
& \textbf{Function class}
& \textbf{Constraint}
& \textbf{Bandit regret}
\\
\midrule
\cite{streeter2008online}
& Adv. & Sub. & Card.
& \(\widetilde{\mathcal O}\!\left(k^2 n^{1/3}T^{2/3}\right)\)
\\
\cite{streeter2009online}
& Adv. & Sub. & Assign.
& \(\widetilde{\mathcal O}\!\left(n^{2/5}k^{6/5}T^{4/5}\right)\)
\\
\cite{zhang2019online}
& Adv. & DR-S / Sub.
& \makecell{Downward-closed convex /\\ Gen.\ matroid}
& \( \calO ( T^{8/9})\)
\\
\cite{nie2022explore}
& Stoch. & Sub. & Card.
& \(\mathcal O\!\left(
    n^{1/3}k^{4/3}T^{2/3}\sqrt{\log T}
  \right)\)
\\
\cite{niazadeh2023online}
& Adv. & Sub. & Card.
& \(\mathcal O\!\left(
    k n^{2/3}(\log n)^{1/3}T^{2/3}
  \right)\)
\\
\cite{wan2023bandit}
& Adv. & Sub. & Part.
& \(\mathcal O\!\left(
    \left(\sum_j r_j|G_j|\right)^{5/3}
    T^{2/3}\log T
  \right)\)
\\
\cite{pasteris2023sum}
& Adv. & Sum-max & Card.
& \(\widetilde{\mathcal O}\!\left(\sqrt{knT}\right)\)
\\
\cite{tajdini2024nearly}
& Stoch. & Sub. & Card.
& \(\widetilde{\mathcal O}\!\left(
    \min_{0\leq L\leq k}
    \left[
      Ln^{1/3}T^{2/3}
      + \sqrt{\binom{n}{k-L}T}
    \right]
  \right)\)
\\
\cite{wan2026bandit}
& Adv. & Sub. & Gen.\ matroid
& \(\widetilde{\mathcal O}\!\left(
    n^{1/3}k^{2/3}T^{2/3}
  \right)\)
\\
\midrule
\rowcolor{gray!15}
\textbf{This work}
& \textbf{Adv.} & \textbf{Sub.} & \textbf{Card.}
& \(\mathcal O\!\left(k\sqrt{nT\log(n/k)}\right)\)
\\
\bottomrule
\end{tabular}%
}
\caption{
Guarantees for $\aReg$ in submodular
maximization with bandit feedback.
Adv.\ and Stoch.\ denote adversarial and stochastic settings;
Sub.\ denotes arbitrary monotone submodular functions;
DR-S denotes monotone DR-submodular functions; and
Card., Assign., Part., and Gen.\ matroid denote cardinality,
assignment, partition-matroid, and general-matroid constraints.
Here, $n$ is the ground-set size and $k$ is the cardinality
budget or matroid rank. For partition matroids, $G_j$ is
the $j$th group and $r_j$ its capacity, so that feasible
sets satisfy $|S\cap G_j|\leq r_j$ for every $j$.
For~\citet{zhang2019online}, the continuous and discrete
settings are listed in that order. Their discrete result
uses responsive-bandit feedback, which allows infeasible
value observations and differs from the strictly feasible
bandit model considered here.
The bound of~\citet{tajdini2024nearly} holds for the stronger
robust-greedy regret and also implies the displayed
upper bound on $\aReg$.
}
\label{tab:bandit-related-work}
\end{table*}

\section{Problem Setting}
\label{sec:setting}

Let \(V=[n]\) be a ground set and let \(2\leq k\leq n\). 
At each round $t\in[T]$, the reward function
$f_t:2^V\to[0,1]$ is monotone and submodular\footnote{The upper bound of $1$ is without loss
of generality: if $f_t(S)\leq B$, $\forall t\in[T], S\subseteq V$, for some known common upper bound $B>0$, we may replace each
$f_t$ by $f_t/B$.}. We assume that $f_t$ is normalized, i.e.,
$f_t(\varnothing)=0$.
The sequence of reward functions $f_1,\ldots,f_T$ is fixed in advance by an oblivious adversary, independently of the learner's random choices.

Formally, a function $f$ is \emph{submodular} iff for all $A \subseteq B$ and $e \in V \setminus B$ it holds that $f(A\cup\{e\}) - f(A) \geq f(B\cup\{e\}) - f(B)$. 
Moreover, $f$ is \emph{monotone} iff for all $A \subseteq B$, it holds $f(A) \leq f(B)$.

On round \(t\), the learner selects a possibly randomized set \(S_t\subseteq V\) satisfying \(|S_t|\leq k\) and observes only the scalar value \(f_t(S_t)\). We measure the performance of the learner via $(1-1/e)$-regret:
\[
\aReg 
\coloneqq
(1-1/e)\max_{\substack{S\subseteq V\\ |S|\leq k}}
\sum_{t=1}^T f_t(S)
-
\E\left[\sum_{t=1}^T f_t(S_t)\right],
\]
where $\E[\cdot]$ is over the learner's random choices.

\spara{Notation.}
Let
$ \Delta_n
    \coloneq
    \left\{
        \bp\in\mathbb{R}_{\geq 0}^n:
        \sum_{i=1}^n p_i=1
    \right\}
$
denote the probability simplex over $V$. Throughout the paper, $\E_t$ denotes
expectation over the learner's sampling randomness on round $t$, conditional
on the preceding history and the selected distribution $\bp_t$.

\section{Full-Information Setting}
\label{sec:full-information}

We begin with the full-information setting, which provides the reduction
underlying our bandit algorithm. 
In this setting, the learner, after playing on round $t$, observes the
reward function $f_t$. 

\subsection{Full-information Algorithm}
The learner uses an online
linear optimization ({\OLO}) algorithm to maintain a distribution
$\bp_t\in\Delta_n$ over the ground set. On each round, it draws $k$ elements
independently from $\bp_t$ with replacement and plays their union $S_t$.
Repeated elements appear only once in $S_t$, so $|S_t|\leq k$.

To describe the linear feedback supplied to the {\OLO} algorithm, we first
define the expected reward of $m$ elements sampled i.i.d. with replacement.
For $\bp\in\Delta_n$, let $B_m$ be the union of $m$ independent draws from
$\bp$, with $B_0\coloneq\varnothing$, and define
\[
F_{t,m}(\bp)\coloneq\E_t[f_t(B_m)].
\]
We extend $F_{t,m}$ to
$\mathbb{R}^n$ through the polynomial
\[
    F_{t,m}(\bp)
    \coloneq \sum_{(i_1,\ldots,i_m)\in V^m}
    f_t(\{i_1,\ldots,i_m\})\prod_{r=1}^m p_{i_r}.
\]
This extension agrees with the sampling expectation on $\Delta_n$;
all gradients below refer to this polynomial. Setting $\rho\coloneq1-1/k$, we
use the potential
\[
    G_t(\bp)
    \coloneq\sum_{m=1}^k\frac{\rho^{k-m}}{m}F_{t,m}(\bp).
\]
The potential combines expected rewards from all draw counts up to $k$.
Its coefficients are chosen so that, for every $\bp\in\Delta_n$ and every set $S\subseteq V$ with $|S|=k$, it holds that
\[
    \left(1-\frac1e\right)f_t(S)-F_{t,k}(\bp)
    \leq
    \left\langle\nabla G_t(\bp),\bp^S-\bp\right\rangle,
\]
where $\bp^S$ is the uniform distribution on $S$.
Thus, the gap between the scaled reward of $S$ and the learner's
expected reward is bounded by the linear reward advantage of $\bp^S$
over $\bp$ under $\nabla G_t(\bp)$.

After observing $f_t$, the learner computes
$\bg_t\coloneq\nabla G_t(\bp_t)$ and supplies the linear reward
$\ell(\bp) \coloneq \langle \bg_t,\bp\rangle$ to the {\OLO} algorithm,
which returns the next distribution $\bp_{t+1}$.
Algorithm~\ref{alg:full-information} summarizes this procedure. In the next subsection, we will prove the following guarantee.
\begin{theorem}[Full-information guarantee, informal]
\label{thm:full-information-informal}
Algorithm~\ref{alg:full-information} transfers the regret bound of its
{\OLO} subroutine to $(1-1/e)$-regret for {\OSM}:
\[
    \aReg\leq\Reg_T^{\mathrm{OLO}}.
\]
In particular, an {\OLO} algorithm with
$\calO(\sqrt{T})$ regret yields $\aReg = \calO(\sqrt{T})$ for {\OSM}.
\end{theorem}

\spara{Discussion.}
This result matches the established $\calO(\sqrt{T})$
$(1-1/e)$-regret rate for full-information {\OSM}; this rate by itself is not a new contribution.
Reducing {\OSM} to {\OLO} is also an established technique, rooted in non-oblivious potentials for offline submodular maximization \citep{filmus2014monotone} and developed through non-oblivious boosting \citep{zhang2022stochastic} and
upper-linearizable functions \citep{pedramfar2024linear}.
Related approaches use non-oblivious potentials under bandit feedback
\citep{wan2023bandit} or reduce {\OSM} to online convex optimization for WTP functions
\citep{salem2024online}.

Our full-information algorithm uses a potential defined through
sampling with replacement. This construction forms the basis of
our bandit algorithm: randomizing the number of draws and choosing
matching potential coefficients allows us to estimate the resulting
potential's gradient without bias from a single observed reward.

We note here that sampling with replacement can, potentially, yield sets with fewer than $k$ distinct elements, leaving part of the cardinality budget unused and potentially limiting practical performance, particularly in the full-information setting.

\begin{algorithm}[t]
\caption{Full-information reduction of {\OSM} to {\OLO}}
\label{alg:full-information}
\begin{algorithmic}[1]
\Require An {\OLO} algorithm $\mathcal{A}$ over $\Delta_n$
\State Set $\rho\gets1-1/k$
\State Initialize $\bp_1 \gets \bone / n$
\For{$t=1,\ldots,T$}
    \State Draw $I_{t,1},\ldots,I_{t,k}
        \overset{\mathrm{i.i.d.}}{\sim}\bp_t$
    \State Play $S_t\gets\{I_{t,1},\ldots,I_{t,k}\}$
    \State Observe $f_t$ and compute
        $\bg_t\gets\nabla G_t(\bp_t)$
    \State Update $\bp_{t+1} \gets \calA(\bg_1, \dots, \bg_t)$ 
\EndFor
\end{algorithmic}
\end{algorithm}

\subsection{Analysis}
\label{subsec:full-information-analysis}

For a set $S\subseteq V$ with $|S|=k$, let
$\bp^S\in\Delta_n$ denote the uniform distribution on $S$. 
By monotonicity, any set with $|S|<k$ can be padded to size $k$ without decreasing its reward.
We first express the gradient of $F_{t,m}$ in terms of the reward obtained
by adding a specified element to $m-1$ independent draws.

\begin{lemma}[Gradient identity]
\label{lem:gradient-identity}
Fix a round $t$, a distribution $\bp\in\Delta_n$, an integer $m\geq1$,
and an element $i\in V$. Let $B_{m-1}$ be the union of $m-1$ independent
draws from $\bp$, with $B_0=\varnothing$. Then
\[
    \frac{\partial F_{t,m}(\bp)}{\partial p_i}
    =m\,\E_t\!\left[f_t(B_{m-1}\cup\{i\})\right].
\]
\end{lemma}

Averaging this identity under $\bp^S$ and $\bp$ and taking the
difference gives the directional derivative from $\bp$ toward $\bp^S$.

\begin{corollary}[Directional derivative identity]
\label{cor:directional-derivative}
Fix a round $t$, a distribution $\bp\in\Delta_n$, an integer $m\geq1$,
and a set $S\subseteq V$ with $|S|=k$. Let $B_{m-1}$ be the union of
$m-1$ independent draws from $\bp$, and let $I_S$ be uniform on $S$,
independently of $B_{m-1}$. Then
\begin{equation*}
    \left\langle\nabla F_{t,m}(\bp),\bp^S-\bp\right\rangle
    =m\left(\E_t\!\left[f_t(B_{m-1}\cup\{I_S\})\right]
    -F_{t,m}(\bp)\right).
\end{equation*}
\end{corollary}

Monotonicity and submodularity give a lower bound on the expected reward
after adding $I_S$ in terms of $f_t(S)$ and the current expected reward.

\begin{lemma}[Submodular averaging]
\label{lem:submodular-averaging}
Fix a round $t$ and let $f_t:2^V\to\mathbb{R}_{\geq0}$ be monotone and
submodular. Let $S\subseteq V$ satisfy $|S|=k$, let $\rho\coloneq1-1/k$,
and let $\bp\in\Delta_n$. For an integer $m\geq1$, let $B_{m-1}$ be
the union of $m-1$ independent draws from $\bp$, and let $I_S$ be uniform
on $S$, independently of $B_{m-1}$. Then
\[
    \E_t\!\left[f_t(B_{m-1}\cup\{I_S\})\right]
    \geq\frac{1}{k}f_t(S)+\rho F_{t,m-1}(\bp).
\]
\end{lemma}

Combining Corollary~\ref{cor:directional-derivative} with
Lemma~\ref{lem:submodular-averaging} gives, for each $m\in[k]$,
\begin{equation*}
    \left\langle\nabla F_{t,m}(\bp),\bp^S-\bp\right\rangle
    \geq m\left(\frac1k f_t(S)
        +\rho F_{t,m-1}(\bp)-F_{t,m}(\bp)\right).
\end{equation*}
Multiplying by $\rho^{k-m}/m$ and summing over $m=1,\ldots,k$
cancels the intermediate terms $F_{t,m}$ for $1\leq m<k$.
Dropping the remaining nonnegative term $\rho^k F_{t,0}$, only the $F_{t,k}$ term survives resulting in the inequality underlying our reduction of {\OSM} to {\OLO}.

\begin{lemma}[Linearization inequality]
\label{lem:full-information-comparator}
Fix a round $t$ and let $f_t:2^V\to\mathbb{R}_{\geq0}$ be monotone and
submodular. Let $S\subseteq V$ satisfy $|S|=k$, let $\bp^S$ be uniform
on $S$, and set $\rho\coloneq1-1/k$ and $\alpha\coloneq1-\rho^k$.
For every $\bp\in\Delta_n$,
\[
    \alpha f_t(S)-F_{t,k}(\bp)
    \leq\left\langle\nabla G_t(\bp),\bp^S-\bp\right\rangle.
\]
\end{lemma}

Since the learner's conditional expected reward satisfies
$\E_t[f_t(S_t)]=F_{t,k}(\bp_t)$ and $\alpha\geq1-1/e$,
summing the linearization inequality over rounds converts regret on linear rewards into $(1-1/e)$-regret for the submodular rewards.
\begin{theorem}[Full-information reduction of {\OSM} to {\OLO}]
\label{thm:full-information}
Let $V=[n]$, let $2\leq k\leq n$, and let
$f_1,\ldots,f_T:2^V\to[0,1]$ be normalized, monotone, submodular reward
functions. Suppose the {\OLO} algorithm in
Algorithm~\ref{alg:full-information} guarantees regret on linear rewards at most
$\Reg_T^{\mathrm{OLO}}$ on the reward vectors
$\{\bg_t=\nabla G_t(\bp_t)\}_{t=1}^T$. Then
\[
    \aReg\leq\Reg_T^{\mathrm{OLO}}.
\]
\end{theorem}

Theorem~\ref{thm:full-information} uses exact gradients. Under value-oracle
access to the revealed function $f_t$, the draws already made by
Algorithm~\ref{alg:full-information} can instead be used to construct
an unbiased gradient estimate, as described in
Appendix~\ref{app:full-info-gradient-estimation}.

Instantiating Theorem~\ref{thm:full-information} with the Exponential Weights algorithm, we get the following corollary.
\begin{corollary}[Full-information regret with the Exponential Weights algorithm]
\label{cor:full-information-regret}
Under the assumptions of Theorem~\ref{thm:full-information},
instantiate the {\OLO} algorithm in Algorithm~\ref{alg:full-information}
with the Exponential Weights algorithm over $\Delta_n$. Then
\[
    \aReg
    =\calO\!\left(k\sqrt{T\log(n/k)}\right).
\]
\end{corollary}

We refer the reader to Appendix~\ref{app:exponential-weights} for more details on the Exponential Weights algorithm.

\section{Bandit Setting}
\label{sec:bandit-algorithm}
\label{sec:bandit-entropy}

In the bandit setting, the learner observes only $f_t(S_t)$ after playing
$S_t$ and therefore cannot compute the potential gradient directly.
The full-information potential $G_t$ contains terms $F_{t,m}$ for every
$m\in[k]$, including $m<k$. Always making $k$ draws reveals only the
reward of their union, without directly revealing the rewards associated
with fewer draws. To estimate all these gradient contributions from a
single reward observation, we randomize the number of draws, allowing
the learner to observe rewards from smaller draw counts as well.

\subsection{Bandit Algorithm}
As in the full-information setting, an {\OLO} algorithm maintains a
distribution $\bp_t\in\Delta_n$. On round $t$, the learner samples a
draw count $M_t\in[k]$ from a fixed distribution
$\bq=(q_1,\ldots,q_k) \in \Delta_k$, draws
$M_t$ elements independently from $\bp_t$ with replacement, and plays
their union $S_t$. Thus, $|S_t|\leq M_t\leq k$.

Using the expected reward $F_{t,m}$ defined in
Section~\ref{sec:full-information}, the learner's expected reward is
\begin{equation}
    H_t(\bp)\coloneq\sum_{m=1}^k q_mF_{t,m}(\bp),
    \qquad \E_t[f_t(S_t)]=H_t(\bp_t),
    \label{eq:bandit-expected-reward}
\end{equation}
where $\E_t$ conditions on the history and the current distribution $\bp_t$.
To account for the random draw count, we use the potential 
\begin{equation}
    G_t(\bp)\coloneq\sum_{m=1}^k a_mF_{t,m}(\bp),
    \label{eq:bandit-potential}
\end{equation}
whose coefficients, with $\rho\coloneq1-1/k$, are
\begin{equation}
    a_m\coloneq\frac{b_m}{m},
    \qquad b_m\coloneq\sum_{\ell=m}^k\rho^{\ell-m}q_\ell.
    \label{eq:bandit-coefficients}
\end{equation}
For $k\geq2$, set $\beta\coloneq1/e-\rho^k$ and
$Z_{k-1}\coloneq\sum_{r=1}^{k-1}r^{-1/2}$. We choose the draw-count
distribution
\begin{equation}
    q_m\coloneq
    \begin{cases}
        \displaystyle\frac{\beta}{Z_{k-1}\sqrt m}, &1\leq m<k,\\[6pt]
        1-\beta, &m=k.
    \end{cases}
    \label{eq:bandit-sampling-distribution}
\end{equation}
Let $N_{t,i}\coloneq\sum_{r=1}^{M_t}\mathbf{1}\{I_{t,r}=i\}$ count the
occurrences of element $i$ among the sampled elements. The observed
reward and these counts give the unbiased gradient estimator
\begin{equation}
    \widetilde{g}_{t,i}
    \coloneq\frac{a_{M_t}f_t(S_t)N_{t,i}}{q_{M_t}p_{t,i}},
    \label{eq:bandit-unbiased-estimator}
\end{equation}
which, for strictly positive $\bp_t$, satisfies
$\E_t[\widetilde{\bg}_t]=\nabla G_t(\bp_t)$.
To preserve the positive probabilities required by its denominator,
we instantiate {\OLO} with exponential weights and use the nonpositive
feedback obtained by subtracting one from the observed reward:
\begin{equation}
    z_{t,i}
    \coloneq\frac{a_{M_t}(f_t(S_t)-1)N_{t,i}}{q_{M_t}p_{t,i}}.
    \label{eq:bandit-feedback}
\end{equation}
Subtracting one shifts the observed reward from $[0,1]$ to $[-1,0]$,
ensuring that every coordinate of $\bz_t$ is nonpositive.
This nonpositivity is used in the exponential-weights analysis below.
The learner supplies the linear reward
$\ell(\bp) = \langle\bz_t,\bp\rangle$ to the {\OLO} algorithm.
Starting from $\bp_1=\mathbf{1}/n$, exponential weights updates the
distribution as
\begin{equation}
    p_{t+1,i}
    =\frac{p_{t,i}\exp(\eta z_{t,i})}
           {\sum_{j=1}^n p_{t,j}\exp(\eta z_{t,j})},
    \label{eq:bandit-entropy-update}
\end{equation}
where $\eta>0$ is the learning rate. The update preserves strictly
positive probabilities, so the feedback in Eq.~\eqref{eq:bandit-feedback}
remains well-defined.
The analysis below shows that the reward shift preserves the expected
linear comparisons needed for the regret reduction.
Algorithm~\ref{alg:bandit} gives the complete procedure.
In the next subsection, we will prove the following result.

\begin{algorithm}[t]
\caption{{\algfull}}
\label{alg:bandit}
\begin{algorithmic}[1]
\Require Ground set $V=[n]$, budget $2\leq k\leq n$,
    horizon $T$, learning rate $\eta>0$
\State $\rho\gets1-1/k$ and $\beta\gets1/e-\rho^k$
\State $Z_{k-1}\gets\sum_{r=1}^{k-1}r^{-1/2}$
\State $q_m\gets\beta/(Z_{k-1}\sqrt m)$ for $m<k$
\State $q_k\gets1-\beta$
\State $b_{k+1}\gets0$
\For{$m=k,k-1,\ldots,1$}
    \State $b_m\gets q_m+\rho b_{m+1}$ and $a_m\gets b_m/m$
\EndFor
\State $\bp_1\gets\mathbf{1}/n$
\For{$t=1,\ldots,T$}
    \State Draw $M_t\sim\bq$
    \State Draw $I_{t,1},\ldots,I_{t,M_t}
        \overset{\mathrm{i.i.d.}}{\sim}\bp_t$
    \State Play $S_t\gets\{I_{t,1},\ldots,I_{t,M_t}\}$
    \State Observe $f_t(S_t)$
    \State $N_{t,i}\gets\sum_{r=1}^{M_t}\mathbf{1}\{I_{t,r}=i\},
        \forall i\in V$
    \State $z_{t,i}\gets
        a_{M_t}(f_t(S_t)-1)N_{t,i}/(q_{M_t}p_{t,i}),\ \forall i\in V$
    \State $\displaystyle p_{t+1,i}\gets
        \frac{p_{t,i}\exp(\eta z_{t,i})}
             {\sum_{j=1}^n p_{t,j}\exp(\eta z_{t,j})},\ \forall i\in V$
\EndFor
\end{algorithmic}
\end{algorithm}

\begin{theorem}[Bandit guarantee, informal]
\label{thm:bandit-informal}
For $2\leq k<n$, against a sequence of normalized,
monotone, submodular reward functions taking values in $[0,1]$,
Algorithm~\ref{alg:bandit} with a suitable learning rate achieves
\[
    \aReg
    =\calO\!\left(k\sqrt{nT\log(n/k)}\right).
\]
\end{theorem}

\spara{Computational Complexity.} Computing $\bq$ and its cumulative probabilities takes
$O(k)$ preprocessing time, as does computing $b_m,a_m$ by the backward
recursion in Algorithm~\ref{alg:bandit}. At each round, constructing
the cumulative probabilities for $\bp_t$ takes $O(n)$ time, sampling at
most $k$ elements by binary search takes $O(k\log n)$ time, and computing
the counts, feedback, and update takes $O(n+k)$ time. Thus, the total running time is
$O(n+k\log n)$ per round and $O(n+k)$ space.

\spara{Online Gradient Descent.}
Appendix~\ref{app:bandit-second-moments} gives an alternative {\OLO}
instantiation using projected Online Gradient Descent (OGD).
This variant operates
over an $\epsilon$-restricted simplex with a positive lower bound on
each coordinate to control the Euclidean second moment. 
The $(1-1/e)$-regret guarantee is of the same $\calO (\sqrt{T})$ order as for Algorithm~\ref{alg:bandit}.

\subsection{Analysis}
\label{subsec:bandit-analysis}

The high-level structure of the analysis is as follows: first, we compare the expected
reward $H_t(\bp_t)$ with a fixed set's reward through the gradient of the potential.
Then, we 
estimate the gradient from the bandit feedback.
Finally, we bound the regret following the analysis of the exponential-weights algorithm (\citep{auer2002nonstochastic}, Appendix~\ref{app:exponential-weights}).

\spara{Bandit linearization inequality.}
Applying Corollary~\ref{cor:directional-derivative} and
Lemma~\ref{lem:submodular-averaging} gives, for each $m\in[k]$,
\begin{equation*}
    \left\langle\nabla F_{t,m}(\bp),\bp^S-\bp\right\rangle
    \geq m\left(\frac1k f_t(S)
        +\rho F_{t,m-1}(\bp)-F_{t,m}(\bp)\right).
\end{equation*}
We multiply by $a_m=b_m/m$ and sum over $m=1,\ldots,k$.
The terms involving
the expected rewards $F_{t, m}$ sum to $\rho b_1F_{t,0}(\bp)-H_t(\bp)$.
Dropping the term $\rho b_1F_{t,0}(\bp)$, which is equal to $0$, gives
\[
    \alpha(\bq)f_t(S)-H_t(\bp)
    \leq\left\langle\nabla G_t(\bp),\bp^S-\bp\right\rangle,
\]
where $\alpha(\bq)\coloneq\frac1k\sum_{m=1}^k b_m
=\sum_{m=1}^k q_m(1-\rho^m)$.
This gives the bandit counterpart of
Lemma~\ref{lem:full-information-comparator}.

\begin{lemma}[Bandit linearization inequality]
\label{lem:bandit-comparator}
Fix a round $t$ and let $f_t:2^V\to\mathbb{R}_{\geq0}$ be monotone and
submodular. For any distribution $\bq \in \Delta_k$, define
$H_t$, $G_t$, and their coefficients by
Eq.~\eqref{eq:bandit-expected-reward}--\eqref{eq:bandit-coefficients}.
For every $\bp\in\Delta_n$ and every $S\subseteq V$ with $|S|=k$,
\[
    \alpha(\bq)f_t(S)-H_t(\bp)
    \leq\langle\nabla G_t(\bp),\bp^S-\bp\rangle.
\]
\end{lemma}

We seek a distribution $\bq$ with
$\alpha(\bq) \geq 1-1/e$. Our bandit gradient estimator also
requires $q_m>0$ for every $m\in[k]$. Concentrating all mass
on $m=k$ recovers the full-information potential and approximation factor
$1-\rho^k$, which exceeds the target factor by $\beta=1/e-\rho^k$.
We use this gap to allocate total mass $\beta$ to draw counts $m < k$
according to Eq.~\eqref{eq:bandit-sampling-distribution}, while retaining
$q_k=1-\beta$. The next lemma verifies that this choice satisfies both
requirements.

\begin{lemma}[Choice of the sampling distribution]
\label{lem:q-choice}
For $k\geq2$, the vector $\bq$ in
Eq.~\eqref{eq:bandit-sampling-distribution} is a probability distribution
with $q_m>0$ for every $m\in[k]$, and
\[
    \alpha(\bq)
    \geq1-\frac1e+\beta\rho^k
    \geq1-\frac1e,
\]
where $\beta = 1/e-\rho^k$.
\end{lemma}
Combining these lemmas gives the comparison used in the regret reduction.

\begin{corollary}[Linearization inequality for the chosen distribution]
\label{cor:bandit-fixed-comparator}
Under the assumptions of Lemma~\ref{lem:bandit-comparator}, choose
$\bq \in \Delta_k$ as in Eq.~\eqref{eq:bandit-sampling-distribution}.
For every $\bp\in\Delta_n$ and every $S\subseteq V$ with $|S|=k$,
\[
    \left(1-\frac1e\right)f_t(S)-H_t(\bp)
    \leq\langle\nabla G_t(\bp),\bp^S-\bp\rangle.
\]
\end{corollary}

\spara{Estimating the potential's gradient from bandit feedback.}
The count $N_{t,i}$ records all positions at which element $i$ was drawn.
Conditional on $M_t=m$, weighting the observed reward by
$N_{t,i}/p_{t,i}$ therefore recovers the derivative in
Lemma~\ref{lem:gradient-identity}. The additional factor $a_m/q_m$
in Eq.~\eqref{eq:bandit-unbiased-estimator} accounts for the potential
coefficient and the probability of selecting that draw count.

\begin{lemma}[Unbiased gradient estimator]
\label{lem:bandit-unbiased-estimator}
Let $\bq \in \Delta_k$ and $\bp_t\in\Delta_n$ have strictly positive
coordinates. Under the sampling rule above,
\[
    \E_t\!\left[
        f_t(S_t)\frac{N_{t,i}}{p_{t,i}}
        \,\middle|\,M_t=m
    \right]
    =\frac{\partial F_{t,m}(\bp_t)}{\partial p_i}.
\]
Consequently,
$\E_t[\widetilde{\bg}_t]=\nabla G_t(\bp_t)$.
\end{lemma}

Subtracting one from the reward makes every coordinate of
$\bz_t$ nonpositive, which permits the exponential-weights
bound below without imposing a uniform lower bound on the sampling
probabilities. Subtracting one from the observed reward also subtracts the same constant
$c_q\coloneq\sum_{m=1}^k b_m$ from every coordinate of the expected feedback.
Thus, the corrected estimator
\begin{equation}
    \widehat{\bg}_t\coloneq\bz_t+c_q\mathbf{1}
    \label{eq:bandit-corrected-estimator}
\end{equation}
is unbiased. The algorithm can use $\bz_t$ directly because
every distribution has total mass one: for $\bu,\bp_t\in\Delta_n$,
\begin{equation}
    \langle\widehat{\bg}_t,\bu-\bp_t\rangle
    =\langle\bz_t+c_q\mathbf{1},\bu-\bp_t\rangle
    =\langle\bz_t,\bu-\bp_t\rangle.
\end{equation}

\begin{lemma}[Expected linear feedback]
\label{lem:estimator-mean}
Under the assumptions of Lemma~\ref{lem:bandit-unbiased-estimator},
the feedback in Eq.~\eqref{eq:bandit-feedback} satisfies
$\E_t[\bz_t]=\nabla G_t(\bp_t)-c_q\mathbf{1}$, and hence
$\E_t[\widehat{\bg}_t]=\nabla G_t(\bp_t)$.
For every fixed $\bu\in\Delta_n$,
\[
    \E_t[\langle\bz_t,\bu-\bp_t\rangle]
    =\langle\nabla G_t(\bp_t),\bu-\bp_t\rangle.
\]
\end{lemma}

Corollary~\ref{cor:bandit-fixed-comparator} compares the submodular
rewards through the potential gradient, and
Lemma~\ref{lem:estimator-mean} replaces that gradient by the observed
feedback in expectation. As in the full-information setting, it is
enough to bound regret on linear rewards against distributions $\bp^S$ that are
uniform on sets of size $k$.

\begin{theorem}[Bandit reduction of {\OSM} to {\OLO}]
\label{thm:bandit-olo}
Let $2\leq k\leq n$, and let $f_1,\ldots,f_T:2^V\to[0,1]$ be
normalized, monotone, submodular functions. Choose $\bq$ as in
Eq.~\eqref{eq:bandit-sampling-distribution} and use the sampling rule and
feedback Eq.~\eqref{eq:bandit-feedback} above. 
Suppose the {\OLO} algorithm
produces strictly positive $\bp_t\in\Delta_n$ and guarantees regret on linear rewards at most $\Reg_T^{\mathrm{OLO}}$ on the reward vectors $\{\bz_t\}_{t=1}^T$. Then,
\[
    \aReg\leq\Reg_T^{\mathrm{OLO}}.
\]
\end{theorem}

\spara{Regret of exponential weights.}
To apply the reduction above, we bound the regret of exponential
weights on the linear rewards defined by $\bz_t$.
The bound requires control of two quantities: the expected weighted
second moment of the feedback and the initial Kullback--Leibler
divergence $D_{\mathrm{KL}}(\bp^S\|\bp_1)$.

We first bound the conditional expectation of
$\sum_i p_{t,i}z_{t,i}^2$. Although the importance weights in
$\bz_t$ can produce large coordinates when some $p_{t,i}$
is small, this expectation admits a bound independent of $\bp_t$ using
$V_q\coloneq\sum_{m=1}^k b_m^2(n+m-1)/(m q_m)$.
This quantity also explains the allocation of probability among draw
counts below $k$: since $b_m\leq1$ and $n+m-1\leq2n$, their contribution
to $V_q$ is at most $2n\sum_{m<k}1/(m q_m)$.
For fixed total mass $\sum_{m<k}q_m=\beta$, this upper bound is minimized
by $q_m\propto1/\sqrt m$, as chosen in
Eq.~\eqref{eq:bandit-sampling-distribution}.

\begin{lemma}[Weighted second moment]
\label{lem:bandit-weighted-moment}
Let $\bq \in \Delta_k$ and $\bp_t\in\Delta_n$ have strictly positive
coordinates, and let $f_t$ take values in $[0,1]$.
Then from Eq.~\eqref{eq:bandit-feedback},
\begin{equation*}
    \E_t\!\left[\sum_{i=1}^n p_{t,i}z_{t,i}^2\right]
    \leq\sum_{m=1}^k\frac{b_m^2(n+m-1)}{m q_m}=V_q.
\end{equation*}
For the distribution in Eq.~\eqref{eq:bandit-sampling-distribution},
$V_q\leq32e\,nk^2$.
\end{lemma}
The second quantity is the initial Kullback--Leibler divergence
$D_{\mathrm{KL}}(\bp^S\|\bp_1)$. Since
$\bp_1=\mathbf{1}/n$ and $\bp^S$ is uniform on a set of size $k$,
this term is $D_{\mathrm{KL}}(\bp^S\|\bp_1)=\log(n/k)$.
Combining the two terms with Theorem~\ref{thm:bandit-olo} gives the
following guarantee.

\begin{theorem}[Exponential-weights guarantee]
\label{thm:bandit-entropy}
Let $2\leq k<n$, and let $f_1,\ldots,f_T:2^V\to[0,1]$ be normalized,
monotone, submodular functions fixed in advance by an oblivious adversary.
For every $\eta>0$, Algorithm~\ref{alg:bandit} satisfies
\[
    \aReg
    \leq\frac{\log(n/k)}{\eta}+\frac{\eta T V_q}{2}.
\]
In particular, choosing $\eta=\sqrt{2\log(n/k)/(T V_q)}$ gives
$
    \aReg
    \leq\calO\!\left(k\sqrt{nT\log(n/k)}\right).
$
\end{theorem}

We refer the reader to Appendix~\ref{app:exponential-weights} for more details on the Exponential Weights algorithm.

\section{Experiments}
\label{sec:experiments}

We complement our theoretical results  with an empirical evaluation of {\alg}.
The experiments illustrate the algorithm's learning behavior and
practical limitations. 

The details of the experimental setup, i.e.,
the datasets, the experimental parameters, and tuning procedures are described
in Appendix~\ref{app:experiments}.
Our code and data are available on \href{https://github.com/jasonNikolaou/bandit-osm}{GitHub}.

\spara{Datasets.}
We consider three monotone submodular maximization problems.
\Influence is a weighted influence-maximization instance.
\TeamFormation is a roster-selection problem with task-dependent
candidate values and redundancy penalties between candidates with
similar skills.
\Coverage is a weighted coverage problem in which elements cover
features whose importance varies across tasks.

\spara{Algorithms.}
We compare {\alg} (Algorithm~\ref{alg:bandit}) with
\OpaqueGreedy~\citep{streeter2008online},
\BanditIG~\citep{niazadeh2023online},
\MSEThree~\citep{pasteris2023sum}, \BanditMLSM~\citep{wan2023bandit}, and \PoissonBE~\citep{wan2026bandit}.
Note that \MSEThree also maintains one item distribution and samples
$k$ times with replacement. Unlike {\alg}, it estimates the
gradient of the fixed-draw expected reward, and its guarantee
relies on pseudo-concavity.
We also include \Random, which selects a uniformly random set of
exactly $k$ distinct elements at each round.

\spara{Metric.}
In all experiments, we report empirical \emph{average approximate regret}
(AAR). For each evaluation run, we define
\[
\mathrm{AAR}(t)
:= \frac{1}{t}\left[
\left(1-\frac{1}{e}\right)
\max_{S\subseteq V,\; |S|\le k}
\sum_{s=1}^{t} f_s(S)
-
\sum_{s=1}^{t} f_s(S_s)
\right].
\]
The comparator is the best fixed feasible set for the first $t$ rounds.
Lower AAR values indicate better performance.
We plot the pointwise mean over ten independent evaluation runs,
with shaded bands indicating one sample standard deviation above
and below the mean.

\begin{figure}[!t]
    \centering

    \begin{minipage}[t]{0.40\textwidth}
        \centering
        \includegraphics[width=\linewidth]{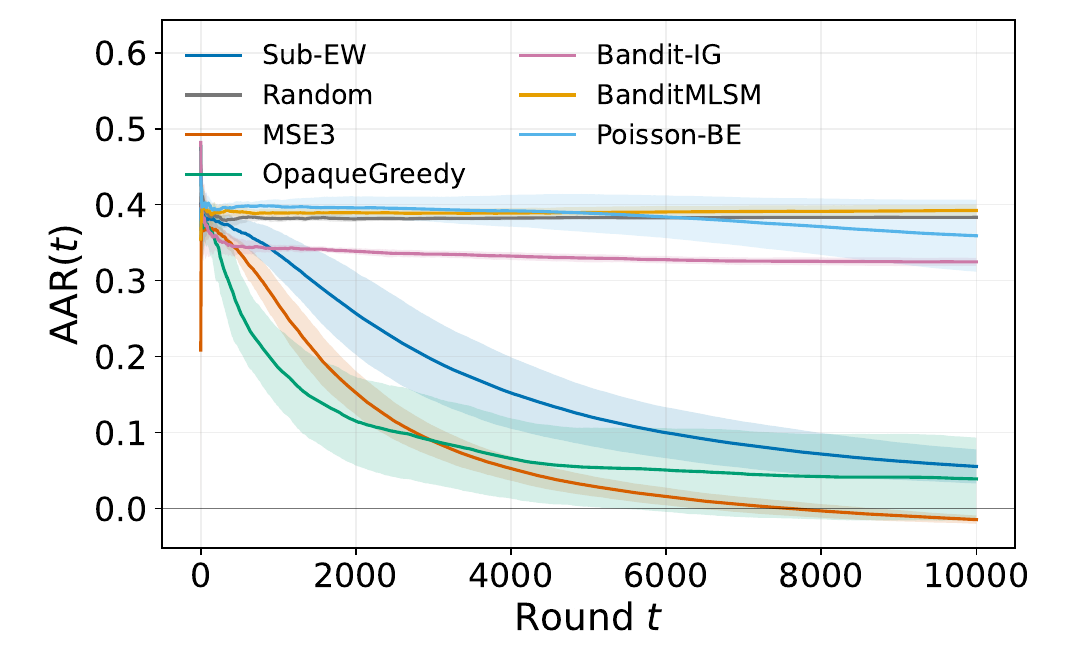}
        \par\smallskip
        (a) \Influence
    \end{minipage}%
    \hspace{0.04\textwidth}%
    \begin{minipage}[t]{0.40\textwidth}
        \centering
        \includegraphics[width=\linewidth]{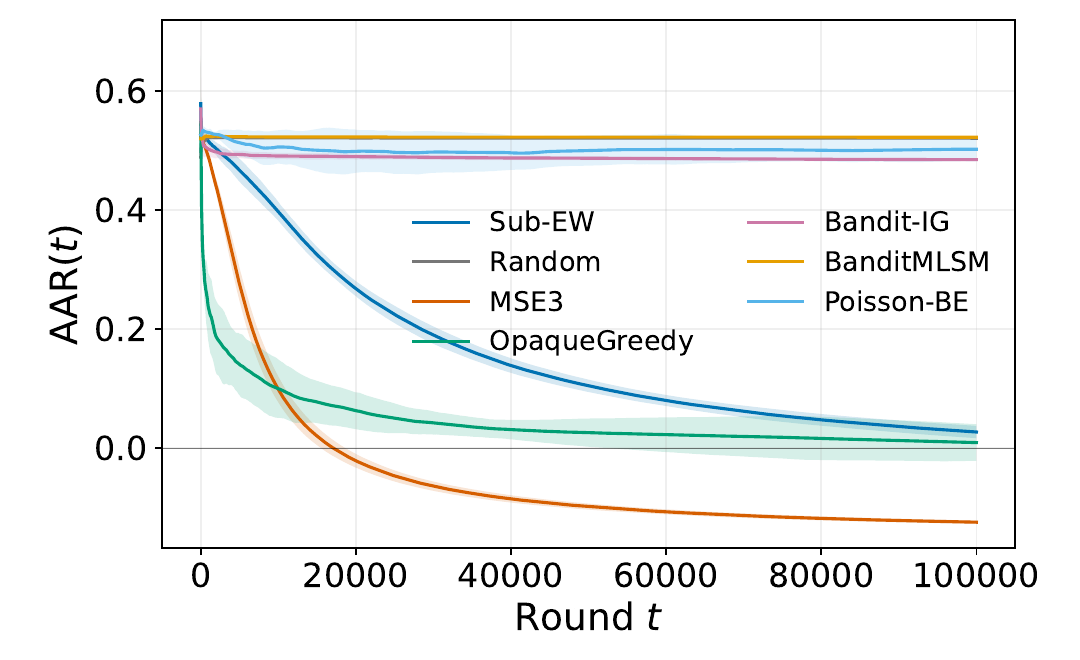}
        \par\smallskip
        (b) \TeamFormation
    \end{minipage}

    \par\medskip

    \begin{minipage}[t]{0.40\textwidth}
        \centering
        \includegraphics[width=\linewidth]{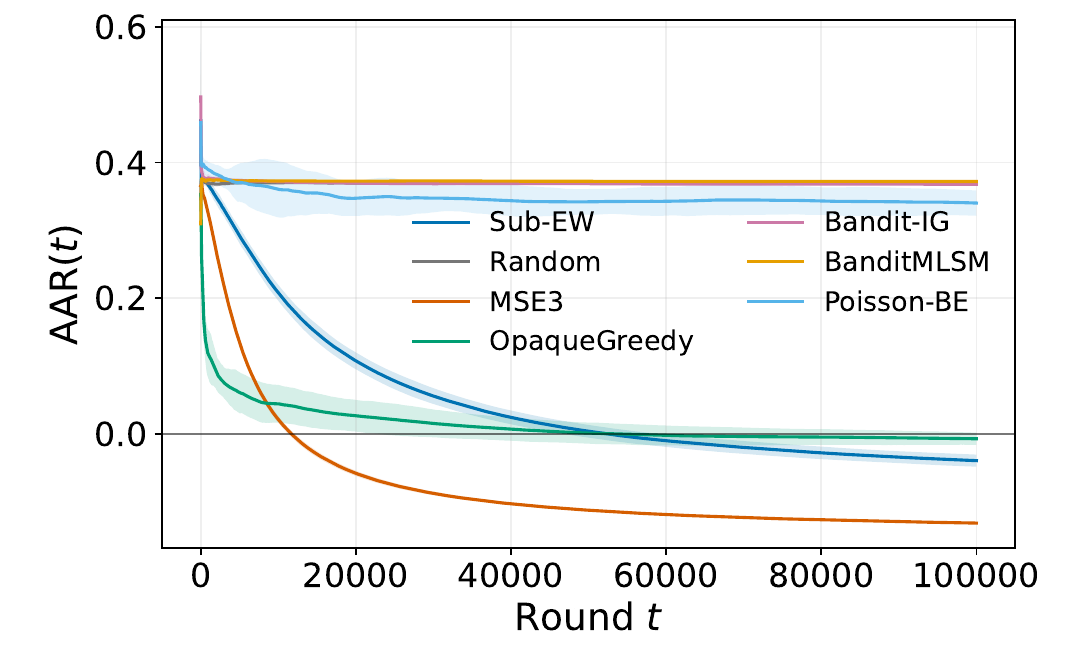}
        \par\smallskip
        (c) \Coverage
    \end{minipage}

    \caption{
        Empirical average approximate regret, $\mathrm{AAR}(t)$, on
        (a) \Influence{} ($n=34$, $k=2$, $T=10^4$),
        (b) \TeamFormation{} ($n=100$, $k=3$, $T=10^5$), and
        (c) \Coverage{} ($n=100$, $k=3$, $T=10^5$).
        Curves show the mean across ten independent evaluation repetitions;
        shaded bands indicate the mean $\pm$ one sample standard deviation.
        Lower values are better.
    }
    \label{fig:experiments}
\end{figure}

\spara{Results.}
Figure~\ref{fig:experiments} shows that {\alg} achieves lower mean final AAR than \Random, \BanditIG, \BanditMLSM, and \PoissonBE across all three datasets.
\MSEThree achieves the lowest mean final AAR on all three datasets.
\OpaqueGreedy also outperforms {\alg} on \Influence and \TeamFormation,
whereas {\alg} performs better on \Coverage.

\spara{Discussion.}
\MSEThree has a $\widetilde{\calO}(\sqrt{T})$ regret guarantee for sum-max functions, which include all three benchmarks, whereas our guarantee holds for arbitrary monotone submodular functions.
\OpaqueGreedy also performs well in practice despite its weaker theoretical regret bound.
Both {\alg} and \MSEThree sample with replacement, potentially leaving part of the cardinality budget unused. However, {\alg} also randomizes the number of draws, sometimes making fewer than $k$, whereas \MSEThree always makes $k$ draws.
This randomization enables estimation of the potential's gradient, but can reduce the obtained reward.
Moreover, the importance-weighted feedback (Eq.~\eqref{eq:bandit-feedback}) can have high variance, potentially slowing learning over finite horizons.

\section{Conclusions}

We achieve $\calO(k\sqrt{nT\log(n/k)})$ expected $(1-1/e)$-regret
for {\problem} under cardinality constraints.
Our approach combines a non-oblivious potential with a randomized number of draws, enabling an unbiased estimate of its gradient from a single observed reward.

\spara{Limitations and Future Work.}
Despite this guarantee, our algorithm trails some baselines in practice.
Designing more effective algorithms with $\sqrt{T}$ regret, and extending
that rate to general matroids, remain open directions.

\spara{AI use statement.}
We used generative AI for reviewing the literature, developing our main research idea of the
potential-function analysis, and improving the presentation throughout the paper. Moreover, we
used it for implementing the experiments in Section 6. Finally, we used it to proofread our work
and pinpoint inconsistencies and minor errors. The authors are responsible for all content of the
paper.

\clearpage
\bibliographystyle{plainnat}
\bibliography{bibliography}

@article{streeter2008online,
  title={An online algorithm for maximizing submodular functions},
  author={Streeter, Matthew and Golovin, Daniel},
  journal={Advances in Neural Information Processing Systems},
  volume={21},
  year={2008}
}

@article{streeter2009online,
  title={Online learning of assignments},
  author={Streeter, Matthew and Golovin, Daniel and Krause, Andreas},
  journal={Advances in neural information processing systems},
  volume={22},
  year={2009}
}

@article{niazadeh2023online,
  title={Online learning via offline greedy algorithms: Applications in market design and optimization},
  author={Niazadeh, Rad and Golrezaei, Negin and Wang, Joshua and Susan, Fransisca and Badanidiyuru, Ashwinkumar},
  journal={Management Science},
  volume={69},
  number={7},
  pages={3797--3817},
  year={2023},
  publisher={INFORMS}
}

@inproceedings{wan2023bandit,
  title={Bandit multi-linear DR-submodular maximization and its applications on adversarial submodular bandits},
  author={Wan, Zongqi and Zhang, Jialin and Chen, Wei and Sun, Xiaoming and Zhang, Zhijie},
  booktitle={International Conference on Machine Learning},
  pages={35491--35524},
  year={2023},
  organization={PMLR}
}

@inproceedings{salem2024online,
  title={Online submodular maximization via online convex optimization},
  author={Si Salem, Tareq and {\"O}zcan, G{\"o}zde and Nikolaou, Iasonas and Terzi, Evimaria and Ioannidis, Stratis},
  booktitle={Proceedings of the AAAI Conference on Artificial Intelligence},
  volume={38},
  pages={15038--15046},
  year={2024}
}

@InProceedings{pasteris2023sum,
  title = 	 {Sum-max Submodular Bandits},
  author =       {Pasteris, Stephen U. and Rumi, Alberto and Vitale, Fabio and Cesa-Bianchi, Nicol\`{o}},
  booktitle = 	 {Proceedings of The 27th International Conference on Artificial Intelligence and Statistics},
  pages = 	 {2323--2331},
  year = 	 {2024},
  editor = 	 {Dasgupta, Sanjoy and Mandt, Stephan and Li, Yingzhen},
  volume = 	 {238},
  series = 	 {Proceedings of Machine Learning Research},
  month = 	 {02--04 May},
  publisher =    {PMLR},
}

@article{wan2026bandit,
  title={Bandit Submodular Maximization under Matroid Constraints: Learning Compressed Exchange Policy},
  author={Wan, Zongqi and Zhang, Zhijie},
  journal={arXiv preprint arXiv:2608.24627},
  year={2026}
}

@inproceedings{nie2022explore,
  title={An explore-then-commit algorithm for submodular maximization under full-bandit feedback},
  author={Nie, Guanyu and Agarwal, Mridul and Umrawal, Abhishek Kumar and Aggarwal, Vaneet and Quinn, Christopher John},
  booktitle={Uncertainty in Artificial Intelligence},
  pages={1541--1551},
  year={2022},
  organization={PMLR}
}

@article{tajdini2024nearly,
  title={Nearly minimax optimal submodular maximization with bandit feedback},
  author={Tajdini, Artin and Jain, Lalit and Jamieson, Kevin},
  journal={Advances in Neural Information Processing Systems},
  volume={37},
  pages={96254--96281},
  year={2024}
}

@article{zhang2019online,
  title={Online continuous submodular maximization: From full-information to bandit feedback},
  author={Zhang, Mingrui and Chen, Lin and Hassani, Hamed and Karbasi, Amin},
  journal={Advances in Neural Information Processing Systems},
  volume={32},
  year={2019}
}

@article{hazan2016introduction,
  title={Introduction to online convex optimization},
  author={Hazan, Elad},
  journal={Foundations and Trends in Optimization},
  volume={2},
  number={3-4},
  pages={157--325},
  year={2016},
  publisher={Emerald Publishing Limited}
}

@article{filmus2014monotone,
  title={Monotone submodular maximization over a matroid via non-oblivious local search},
  author={Filmus, Yuval and Ward, Justin},
  journal={SIAM Journal on Computing},
  volume={43},
  number={2},
  pages={514--542},
  year={2014},
  publisher={SIAM}
}

@inproceedings{zhang2022stochastic,
  title={Stochastic continuous submodular maximization: Boosting via non-oblivious function},
  author={Zhang, Qixin and Deng, Zengde and Chen, Zaiyi and Hu, Haoyuan and Yang, Yu},
  booktitle={International Conference on Machine Learning},
  pages={26116--26134},
  year={2022},
  organization={PMLR}
}

@article{pedramfar2024linear,
  title={From linear to linearizable optimization: A novel framework with applications to stationary and non-stationary dr-submodular optimization},
  author={Pedramfar, Mohammad and Aggarwal, Vaneet},
  journal={Advances in Neural Information Processing Systems},
  volume={37},
  pages={37626--37664},
  year={2024}
}

@inproceedings{kempe2003maximizing,
  title={Maximizing the spread of influence through a social network},
  author={Kempe, David and Kleinberg, Jon and Tardos, {\'E}va},
  booktitle={Proceedings of the ninth ACM SIGKDD international conference on Knowledge discovery and data mining},
  pages={137--146},
  year={2003}
}

@article{zachary1977information,
  title={An information flow model for conflict and fission in small groups},
  author={Zachary, Wayne W},
  journal={Journal of anthropological research},
  volume={33},
  number={4},
  pages={452--473},
  year={1977},
  publisher={University of New Mexico}
}

@inproceedings{krause2007near,
  title={Near-optimal observation selection using submodular functions},
  author={Krause, Andreas and Guestrin, Carlos},
  booktitle={AAAI},
  volume={7},
  pages={1650--1654},
  year={2007}
}

@article{xu2023online,
  title={Online submodular coordination with bounded tracking regret: Theory, algorithm, and applications to multi-robot coordination},
  author={Xu, Zirui and Zhou, Hongyu and Tzoumas, Vasileios},
  journal={IEEE Robotics and Automation Letters},
  volume={8},
  number={4},
  pages={2261--2268},
  year={2023},
  publisher={IEEE}
}

@article{li2021online,
  title={Online caching networks with adversarial guarantees},
  author={Li, Yuanyuan and Si Salem, Tareq and Neglia, Giovanni and Ioannidis, Stratis},
  journal={Proceedings of the ACM on Measurement and Analysis of Computing Systems},
  volume={5},
  number={3},
  pages={1--39},
  year={2021},
  publisher={ACM New York, NY, USA}
}

@article{nikolaou2026online,
  title={Online Two-Stage Submodular Maximization},
  author={Nikolaou, Iasonas and Stouras, Miltiadis and Ioannidis, Stratis and Terzi, Evimaria},
  journal={Advances in Neural Information Processing Systems},
  volume={38},
  pages={2128--2167},
  year={2025}
}

@inproceedings{bhowmik2014submodularity,
  title={Submodularity in team formation problem},
  author={Bhowmik, Avradeep and Borkar, Vivek and Garg, Dinesh and Pallan, Madhavan},
  booktitle={Proceedings of the 2014 SIAM international conference on data mining},
  pages={893--901},
  year={2014},
  organization={SIAM}
}

@article{feige1998threshold,
  title={A threshold of ln n for approximating set cover},
  author={Feige, Uriel},
  journal={Journal of the ACM (JACM)},
  volume={45},
  number={4},
  pages={634--652},
  year={1998},
  publisher={ACM New York, NY, USA}
}

@article{auer2002nonstochastic,
  title={The nonstochastic multiarmed bandit problem},
  author={Auer, Peter and Cesa-Bianchi, Nicolo and Freund, Yoav and Schapire, Robert E},
  journal={SIAM journal on computing},
  volume={32},
  number={1},
  pages={48--77},
  year={2002},
  publisher={SIAM}
}

\clearpage
\appendix
\onecolumn
\thispagestyle{empty}

\section{Full-information algorithm proofs}
\label{app:full-info}

\subsection{Proof of Lemma~\ref{lem:gradient-identity}}
\label{app:gradient-identity}
\begin{restatedlemma}[Restatement of Lemma~\ref{lem:gradient-identity}, Gradient identity]
Fix a round $t$, a distribution $\bp\in\Delta_n$, an integer
$m\geq 1$, and an element $i\in V$. Let $B_{m-1}$ be the union of
$m-1$ independent draws from $\bp$, with $B_0=\varnothing$.
Then
\[
    \frac{\partial F_{t,m}(\bp)}{\partial p_i}
    =
    m\,\E\!\left[f_t(B_{m-1}\cup\{i\})\right].
\]
\end{restatedlemma}

\begin{proof}
Differentiating the polynomial definition gives
\[
\begin{aligned}
    \frac{\partial F_{t,m}(\bp)}{\partial p_i}
    &=
    \sum_{j_1,\ldots,j_m\in V}
    f_t(\{j_1,\ldots,j_m\})
    \frac{\partial}{\partial p_i}
    \left(\prod_{r=1}^m p_{j_r}\right)\\
    &=
    \sum_{s=1}^m
    \sum_{(j_r)_{r\neq s}\in V^{m-1}}
    \left(\prod_{r\neq s}p_{j_r}\right)
    f_t\!\left(\{i\}\cup\{j_r:r\neq s\}\right).
\end{aligned}
\]
For each fixed $s$, the inner sum is
$\E[f_t(B_{m-1}\cup\{i\})]$: the draw in position $s$ is fixed to $i$,
while the other $m-1$ draws are independent samples from
$\bp$. All $m$ positions therefore make the same contribution,
which proves the identity.
\end{proof}

\subsection{Proof of Corollary~\ref{cor:directional-derivative}}
\label{app:gradient-comparison-corollary}

\begin{restatedcorollary}[Restatement of Corollary~\ref{cor:directional-derivative}, Directional derivative identity]
Fix a round $t$, a distribution $\bp\in\Delta_n$, an integer $m\geq1$,
and a set $S\subseteq V$ with $|S|=k$. Let $B_{m-1}$ be the union of
$m-1$ independent draws from $\bp$, and let $I_S$ be uniform on $S$,
independently of $B_{m-1}$. Then
\[
    \left\langle\nabla F_{t,m}(\bp),\bp^S-\bp\right\rangle
    \\
    =m\left(\E_t\!\left[f_t(B_{m-1}\cup\{I_S\})\right]
    -F_{t,m}(\bp)\right).
\]
\end{restatedcorollary}

\begin{proof}
By Lemma~\ref{lem:gradient-identity} and the definition of
$\bp^S$,
\[
\begin{aligned}
    \left\langle
        \nabla F_{t,m}(\bp),\bp^S
    \right\rangle
    &=
    \frac{m}{k}\sum_{i\in S}
    \E\!\left[f_t(B_{m-1}\cup\{i\})\right]\\
    &=
    m\,\E\!\left[f_t(B_{m-1}\cup\{I_S\})\right].
\end{aligned}
\]
Similarly, if $I\sim\bp$ independently of $B_{m-1}$, then
\[
\begin{aligned}
    \left\langle
        \nabla F_{t,m}(\bp),\bp
    \right\rangle
    &=
    m\sum_{i\in V}p_i
    \E\!\left[f_t(B_{m-1}\cup\{i\})\right]\\
    &=
    m\,\E\!\left[f_t(B_{m-1}\cup\{I\})\right] \\
    &=mF_{t,m}(\bp),
\end{aligned}
\]
since $B_{m-1}\cup\{I\}$ has the same distribution as $B_m$.
Subtracting the two identities proves the claim.
\end{proof}

\subsection{Proof of Lemma~\ref{lem:submodular-averaging}}
\label{app:submodular-averaging}

\begin{restatedlemma}[Restatement of Lemma~\ref{lem:submodular-averaging}, Submodular averaging]
Fix a round $t$ and let $f_t:2^V\to\mathbb{R}_{\geq 0}$ be monotone and
submodular. Let $S\subseteq V$ satisfy $|S|=k$, let
$\rho\coloneq1-1/k$, and let $\bp\in\Delta_n$. For an integer
$m\geq 1$, let $B_{m-1}$ be the union of $m-1$ independent draws from
$\bp$, and let $I_S$ be sampled uniformly from $S$, independently
of $B_{m-1}$. Then
\[
    \E\!\left[f_t(B_{m-1}\cup\{I_S\})\right]
    \geq
    \frac{1}{k}f_t(S)
    +\rho F_{t,m-1}(\bp).
\]
\end{restatedlemma}

\begin{proof}
Fix $B\subseteq V$. Submodularity implies
\[
    \sum_{i\in S}
    \bigl(f_t(B\cup\{i\})-f_t(B)\bigr)
    \geq
    f_t(B\cup S)-f_t(B).
\]
Therefore, by monotonicity,
\[
\begin{aligned}
    \frac{1}{k}\sum_{i\in S}f_t(B\cup\{i\})
    &\geq
    \left(1-\frac{1}{k}\right)f_t(B)
    +\frac{1}{k}f_t(B\cup S)\\
    &\geq
    \rho f_t(B)+\frac{1}{k}f_t(S).
\end{aligned}
\]
Apply this inequality to $B=B_{m-1}$ and take expectation over
$B_{m-1}$ and the independent uniform element $I_S$. This yields
\[
\begin{aligned}
    \E\!\left[f_t(B_{m-1}\cup\{I_S\})\right]
    &=
    \E\!\left[
        \frac{1}{k}\sum_{i\in S}f_t(B_{m-1}\cup\{i\})
    \right]\\
    &\geq
    \frac{1}{k}f_t(S)
    +\rho\,\E[f_t(B_{m-1})]\\
    &=
    \frac{1}{k}f_t(S)
    +\rho F_{t,m-1}(\bp).
\end{aligned}
\]
\end{proof}

\subsection{Proof of Lemma~\ref{lem:full-information-comparator}}
\label{app:full-information-comparator}

\begin{restatedlemma}[Restatement of Lemma~\ref{lem:full-information-comparator}, Linearization inequality]
Fix a round $t$ and let $f_t:2^V\to\mathbb{R}_{\geq 0}$ be monotone and submodular. Let $S\subseteq V$ satisfy $|S|=k$, let $\bp^S\in\Delta_n$ be the uniform distribution on $S$, and set $\rho\coloneq1-1/k$. Then, for every $\bp\in\Delta_n$,
\[
    \bigl(1-\rho^k\bigr)f_t(S)-F_{t,k}(\bp)
    \leq
    \left\langle
        \nabla G_t(\bp),
        \bp^S-\bp
    \right\rangle.
\]
\end{restatedlemma}

\begin{proof}
By Corollary~\ref{cor:directional-derivative} and
Lemma~\ref{lem:submodular-averaging}, for each $m\in[k]$,
\[
\begin{aligned}
    \left\langle
        \nabla F_{t,m}(\bp),
        \bp^S-\bp
    \right\rangle
    &\geq
    \frac{m}{k}f_t(S)
    +m\rho F_{t,m-1}(\bp)
    -mF_{t,m}(\bp).
\end{aligned}
\]
Multiplying by $\rho^{k-m}/m$ and summing over $m=1,\ldots,k$ gives
\[
\begin{aligned}
    \left\langle
        \nabla G_t(\bp),
        \bp^S-\bp
    \right\rangle
    \geq {}&
    \frac{1}{k}\sum_{m=1}^k\rho^{k-m}f_t(S)\\
    &+
    \sum_{m=1}^k\rho^{k-m+1}F_{t,m-1}(\bp)
    -
    \sum_{m=1}^k\rho^{k-m}F_{t,m}(\bp).
\end{aligned}
\]
Because $F_{t,0}(\bp)=0$, the last two sums telescope to
$-F_{t,k}(\bp)$. Moreover,
\[
    \frac{1}{k}\sum_{m=1}^k\rho^{k-m}
    =
    \frac{1}{k}\sum_{\ell=0}^{k-1}\rho^\ell
    =
    1-\rho^k,
\]
where the last equality uses $1-\rho=1/k$. Substituting these two identities
proves the claim.
\end{proof}

\subsection{Proof of Theorem~\ref{thm:full-information}}
\label{app:thm-full-information}

\begin{restatedtheorem}[Restatement of Theorem~\ref{thm:full-information}, Full-information reduction of {\OSM} to {\OLO}]
Let $V=[n]$, let $2\leq k\leq n$, and let
$f_1,\ldots,f_T:2^V\to[0,1]$ be normalized, monotone, submodular reward
functions.
Suppose Algorithm~\ref{alg:full-information} uses an {\OLO}
algorithm that for every sequence
of linear reward vectors $\bg_1,\ldots,\bg_T$, guarantees regret $\Reg_T^{\mathrm{OLO}}$.
When the {\OLO} algorithm receives
$\bg_t=\nabla G_t(\bp_t)$ on every round,
Algorithm~\ref{alg:full-information} satisfies
\[
    \aReg
    \coloneq
    \left(1-\frac{1}{e}\right)
    \max_{S\subseteq V:\,|S|\leq k}
    \sum_{t=1}^T f_t(S)
    -
    \E\!\left[\sum_{t=1}^T f_t(S_t)\right]
    \leq
    \Reg_T^{\mathrm{OLO}}.
\]
\end{restatedtheorem}

\begin{proof}
Set $\rho \coloneq 1-1/k$ and $\alpha \coloneq 1-\rho^k$. Fix any
$S\subseteq V$ with $|S|=k$. Lemma~\ref{lem:full-information-comparator}
implies, on every round $t$,
\[
    \alpha f_t(S)-F_{t,k}(\bp_t)
    \leq
    \left\langle
        \nabla G_t(\bp_t),
        \bp^S-\bp_t
    \right\rangle.
\]
Summing over $t$ and applying the {\OLO} guarantee with
$\bu=\bp^S$ and
$\bg_t=\nabla G_t(\bp_t)$ yields
\[
    \alpha\sum_{t=1}^T f_t(S)
    -\sum_{t=1}^T F_{t,k}(\bp_t)
    \leq
    \Reg_T^{\mathrm{OLO}}.
\]
Conditional on the history and $\bp_t$, the set $S_t$ is the union
of $k$ independent draws from $\bp_t$. Hence
$\E_t[f_t(S_t)]=F_{t,k}(\bp_t)$. Taking expectations and using the
tower property therefore gives
\[
    \alpha\sum_{t=1}^T f_t(S)
    -\E\!\left[\sum_{t=1}^T f_t(S_t)\right]
    \leq
    \Reg_T^{\mathrm{OLO}}.
\]
Maximizing over all $S$ with $|S|=k$ and using
$\alpha=1-(1-1/k)^k\geq 1-1/e$ concludes the proof.
\end{proof}

\subsection{Proof of Corollary~\ref{cor:full-information-regret}}
\label{app:cor-full-information-regret}

\begin{restatedcorollary}[Restatement of Corollary~\ref{cor:full-information-regret}, Full-information regret with the Exponential Weights algorithm]
Under the assumptions of Theorem~\ref{thm:full-information},
instantiate the {\OLO} algorithm in Algorithm~\ref{alg:full-information}
with the Exponential Weights algorithm over $\Delta_n$. Then
\[
    \aReg
    =\calO\!\left(k\sqrt{T\log(n/k)}\right).
\]
\end{restatedcorollary}

\begin{proof}
Set $\rho:=1-1/k$. By the gradient identity for $F_{t,m}$ and the
definition of $G_t$, each coordinate of $\bg_t=\nabla G_t(\bp_t)$ satisfies
\[
    g_{t,i}
    =\sum_{m=1}^k \rho^{k-m}
      \E\!\left[f_t(B_{t,m-1}\cup\{i\})\mid\bp_t\right]
    \in[0,k],
\]
where $B_{t,m-1}$ is the union of $m-1$ independent draws from $\bp_t$.
Fix $S\subseteq V$ with $|S|=k$, and let $\bp^S$ be uniform on $S$.
Exponential Weights starts from $\bp_1=(1/n,\ldots,1/n)$, so
\[
    D_{\mathrm{KL}}(\bp^S\|\bp_1)=\log(n/k).
\]
For any $\eta>0$, its full-information regret bound
(Appendix~\ref{app:ew-full-information}) therefore gives
\[
    \sum_{t=1}^T\langle\bg_t,\bp^S-\bp_t\rangle
    \leq \frac{\log(n/k)}{\eta}+\frac{\eta k^2T}{8}.
\]
Set $\alpha:=1-\rho^k$. The comparator inequality used in the proof of
Theorem~\ref{thm:full-information} gives
\[
    \alpha\sum_{t=1}^T f_t(S)
    -\sum_{t=1}^T F_{t,k}(\bp_t)
    \leq \sum_{t=1}^T\langle\bg_t,\bp^S-\bp_t\rangle
    \leq \frac{\log(n/k)}{\eta}+\frac{\eta k^2T}{8}.
\]
Since $\E_t[f_t(S_t)]=F_{t,k}(\bp_t)$ and
$\alpha\geq1-1/e$, taking expectations and maximizing over $S$ yields
\[
    \aReg\leq\frac{\log(n/k)}{\eta}+\frac{\eta k^2T}{8}.
\]
Choose
$\eta=\sqrt{8\log(n/k)/(k^2T)}$ to obtain
\[
    \aReg\leq k\sqrt{\frac{T\log(n/k)}{2}}.
\]
\end{proof}

\subsection{Gradient estimation}
\label{app:full-info-gradient-estimation}
Under value-oracle access to the revealed function $f_t$, one can
instead construct an unbiased gradient estimate from the draws already made by Algorithm~\ref{alg:full-information}. 
Let
$B_{t,0}\coloneq\varnothing$ and
$B_{t,r}\coloneq\{I_{t,1},\ldots,I_{t,r}\}$ for $r=1,\ldots,k-1$, and define
$\widetilde{\bg}_t \in\mathbb{R}^n$ coordinatewise by
\[
    \widetilde{g}_{t,i}
    \coloneq
    \sum_{m=1}^k
    \rho^{k-m}f_t(B_{t,m-1}\cup\{i\}).
\]
Conditional on the preceding history and $\bp_t$, each
$B_{t,m-1}$ is the union of $m-1$ independent draws from
$\bp_t$. Lemma~\ref{lem:gradient-identity} therefore gives
\[
\begin{aligned}
    \E_t[\widetilde{g}_{t,i}]
    &=
    \sum_{m=1}^k\rho^{k-m}
    \E_t[f_t(B_{t,m-1}\cup\{i\})]\\
    &=
    \sum_{m=1}^k
    \frac{\rho^{k-m}}{m}
    \frac{\partial F_{t,m}(\bp_t)}{\partial p_i}\\
    &=
    \frac{\partial G_t(\bp_t)}{\partial p_i}.
\end{aligned}
\]
The prefixes need not be independent across $m$, because this calculation
uses only linearity of expectation. Moreover,
$0\leq\widetilde{g}_{t,i}\leq k$, and constructing
$\widetilde{\bg}_t$ requires at most $nk$ value queries per round.
Thus, an {\OLO} algorithm with an expected regret guarantee on linear rewards may receive
$\widetilde{\bg}_t$ in place of the exact gradient, and the same
regret-transfer argument applies in expectation.
\section{Bandit feedback algorithm proofs}
\label{app:bandit}

\subsection{Proof of Lemma~\ref{lem:bandit-comparator}}
\begin{restatedlemma}[Restatement of Lemma~\ref{lem:bandit-comparator}, Bandit linearization inequality]
Fix a round $t$ and let $f_t:2^V\to\mathbb{R}_{\geq0}$ be monotone and
submodular. For any distribution $\bq \in \Delta_k$, define
$H_t$, $G_t$, and their coefficients by
Eq.~\eqref{eq:bandit-expected-reward}--\eqref{eq:bandit-coefficients}.
For every $\bp\in\Delta_n$ and every $S\subseteq V$ with $|S|=k$,
\[
    \alpha(\bq)f_t(S)-H_t(\bp)
    \leq\langle\nabla G_t(\bp),\bp^S-\bp\rangle.
\]
\end{restatedlemma}

\begin{proof}
The gradient identity and submodular averaging inequality from
the full-information analysis give, for each $m\in[k]$,
\[
    \langle\nabla F_{t,m}(\bp),\bp^S-\bp\rangle
    \geq
    \frac{m}{k}f_t(S)+m\rho F_{t,m-1}(\bp)-mF_{t,m}(\bp).
\]
Multiplying by $b_m/m$ and summing over $m$ yields
\[
    \langle\nabla G_t(\bp),\bp^S-\bp\rangle
    \geq
    \frac1k\sum_{m=1}^k b_m f_t(S)
    +\sum_{m=1}^k b_m
        \bigl(\rho F_{t,m-1}(\bp)-F_{t,m}(\bp)\bigr).
\]
Using $F_{t,0}=0$, $b_{k+1}=0$, and
$b_m=q_m+\rho b_{m+1}$, the last sum is
\[
    \sum_{m=1}^k(\rho b_{m+1}-b_m)F_{t,m}(\bp)
    =-H_t(\bp).
\]
Also, changing the order of summation gives
\[
    \frac1k\sum_{m=1}^k b_m
    =
    \sum_{\ell=1}^k q_\ell
    \left(\frac1k\sum_{m=1}^{\ell}\rho^{\ell-m}\right)
    =
    \sum_{\ell=1}^k q_\ell(1-\rho^\ell)
    =\alpha(\bq).
\]
Substituting these identities proves the claim.
\end{proof}

\subsection{Proof of Corollary~\ref{cor:bandit-fixed-comparator}}

\begin{restatedcorollary}[Restatement of Corollary~\ref{cor:bandit-fixed-comparator}, Linearization inequality for the chosen distribution]
Under the assumptions of Lemma~\ref{lem:bandit-comparator}, choose
$\bq \in \Delta_k$ as in Eq.~\eqref{eq:bandit-sampling-distribution}.
For every $\bp\in\Delta_n$ and every $S\subseteq V$ with $|S|=k$,
\[
    \left(1-\frac1e\right)f_t(S)-H_t(\bp)
    \leq\langle\nabla G_t(\bp),\bp^S-\bp\rangle.
\]
\end{restatedcorollary}

\begin{proof}
Apply Lemmas~\ref{lem:bandit-comparator} and~\ref{lem:q-choice},
using $f_t(S)\geq0$.
\end{proof}

\subsection{Proof of Lemma~\ref{lem:bandit-unbiased-estimator}}

\begin{restatedlemma}[Restatement of Lemma~\ref{lem:bandit-unbiased-estimator}, Unbiased gradient estimator]
Let $\bq \in \Delta_k$ have strictly positive coordinates, and
let $\bp_t\in\Delta_n$ be strictly positive.
Under the sampling rule above,
\[
    \E_t\!\left[
        f_t(S_t)\frac{N_{t,i}}{p_{t,i}}
        \,\middle|\,M_t=m
    \right]
    =\frac{\partial F_{t,m}(\bp_t)}{\partial p_i}.
\]
Consequently,
$
    \E_t[\widetilde{\bg}_t]
    =\nabla G_t(\bp_t).
$
\end{restatedlemma}

\begin{proof}
Expanding $N_{t,i}$ and conditioning on each draw gives
\begin{align*}
    \E_t\!\left[
        f_t(S_t)\frac{N_{t,i}}{p_{t,i}}
        \,\middle|\,M_t=m
    \right]
    &=\sum_{r=1}^m
      \E_t[f_t(S_t)\mid M_t=m,\ I_{t,r}=i]\\
    &=m\,\E[f_t(B_{m-1}\cup\{i\})]\\
    &=\frac{\partial F_{t,m}(\bp_t)}{\partial p_i},
\end{align*}
where $B_{m-1}$ is the union of the remaining $m-1$ independent
draws from $\bp_t$. The last equality is
Lemma~\ref{lem:gradient-identity}.
Multiplying by $a_m/q_m$ and averaging over $M_t$ proves
the second claim.
\end{proof}

\subsection{Proof of Lemma~\ref{lem:estimator-mean}}
\begin{restatedlemma}[Restatement of Lemma~\ref{lem:estimator-mean}, Expected linear feedback]
Under the assumptions of Lemma~\ref{lem:bandit-unbiased-estimator},
the feedback in Eq.~\eqref{eq:bandit-feedback} satisfies
$\E_t[\bz_t]=\nabla G_t(\bp_t)-c_q\mathbf{1}$, and hence
$\E_t[\widehat{\bg}_t]=\nabla G_t(\bp_t)$.
For every fixed $\bu\in\Delta_n$,
\[
    \E_t[\langle\bz_t,\bu-\bp_t\rangle]
    =\langle\nabla G_t(\bp_t),\bu-\bp_t\rangle.
\]
\end{restatedlemma}

\begin{proof}
Since $\E_t[N_{t,i}\mid M_t=m]=mp_{t,i}$,
\[
    \E_t[\widetilde{g}_{t,i}-z_{t,i}]
    =\sum_{m=1}^k a_m m=c_q.
\]
The first claim follows from
Lemma~\ref{lem:bandit-unbiased-estimator}. The second uses
$\langle\mathbf{1},\bu-\bp_t\rangle=0$.
\end{proof}

\subsection{Proof of Lemma~\ref{lem:q-choice}}
\begin{restatedlemma}[Restatement of Lemma~\ref{lem:q-choice}, Choice of the sampling distribution]
Let $k\geq2$, $\rho\coloneq1-1/k$,
$\beta\coloneq1/e-\rho^k$, $Z_{k-1}\coloneq\sum_{r=1}^{k-1}r^{-1/2}$ and define
\[
    q_m\coloneq
    \begin{cases}
        \displaystyle\frac{\beta}{Z_{k-1}\sqrt m},
            &1\leq m<k,\\[6pt]
        1-\beta, &m=k.
    \end{cases}
\]
Then $\bq$ is a probability distribution with
$q_m>0$ for every $m\in[k]$, and
\[
    \alpha(\bq)
    \coloneq\sum_{m=1}^k q_m(1-\rho^m)
    \geq 1 - 1/e + \beta \rho^k
    \geq1-\frac1e.
\]
\end{restatedlemma}

\begin{proof}
Since $k\geq2$ and $\rho=1-1/k$, we have
$0<\rho^k<1/e$ and hence $0<\beta<1$.
Every $q_m$ is positive, and
\[
    \sum_{m=1}^{k-1}q_m
    =
    \frac{\beta}{Z_{k-1}}
    \sum_{m=1}^{k-1}\frac{1}{\sqrt{m}}
    =\beta.
\]
Together with $q_k=1-\beta$, this shows that the probabilities
sum to one. Since every $1-\rho^m$ is nonnegative,
\[
    \alpha(\bq)
    =
    \sum_{m=1}^k q_m(1-\rho^m)
    \geq
    (1-\beta)(1-\rho^k)
    =
    1-\frac1e+\beta\rho^k
    \geq1-\frac1e,
\]
where the equality uses $\beta+\rho^k=1/e$.
\end{proof}

\subsection{Proof of Theorem~\ref{thm:bandit-olo}}

\begin{restatedtheorem}[Restatement of Theorem~\ref{thm:bandit-olo}, Bandit reduction of {\OSM} to {\OLO}]
Let $2\leq k\leq n$, and let $f_1,\ldots,f_T:2^V\to[0,1]$ be
normalized, monotone, submodular functions. Choose $\bq$ as in
Eq.~\eqref{eq:bandit-sampling-distribution} and use the sampling rule and
feedback Eq.~\eqref{eq:bandit-feedback} above. 
Suppose the {\OLO} algorithm
produces strictly positive $\bp_t\in\Delta_n$ and guarantees regret on linear rewards at most $\Reg_T^{\mathrm{OLO}}$ on the reward vectors $\{\bz_t\}_{t=1}^T$. Then,
\[
    \aReg\leq\Reg_T^{\mathrm{OLO}}.
\]
\end{restatedtheorem}

\begin{proof}
Fix $S\subseteq V$ with $|S|=k$.
By Corollary~\ref{cor:bandit-fixed-comparator},
$\E_t[f_t(S_t)]=H_t(\bp_t)$, and
Lemma~\ref{lem:estimator-mean},
\[
    \left(1-\frac1e\right)f_t(S)-\E_t[f_t(S_t)]
    \leq
    \E_t[\langle\bz_t,\bp^S-\bp_t\rangle].
\]
Taking expectations, summing over $t$, and applying the {\OLO}
guarantee gives
\[
    \left(1-\frac1e\right)\sum_{t=1}^T f_t(S)
    -
    \E\!\left[\sum_{t=1}^T f_t(S_t)\right]
    \leq\Reg_T^{\mathrm{OLO}}.
\]
Maximizing over $S$ concludes the proof.
\end{proof}

\subsection{Proof of Lemma~\ref{lem:bandit-weighted-moment}}

\begin{restatedlemma}[Restatement of Lemma~\ref{lem:bandit-weighted-moment}, Weighted second moment]
Let $\bq \in \Delta_k$ and $\bp_t\in\Delta_n$ have strictly positive
coordinates, and let $f_t$ take values in $[0,1]$.
Then from Eq.~\eqref{eq:bandit-feedback},
\[    \E_t\!\left[\sum_{i=1}^n p_{t,i}z_{t,i}^2\right]
    \leq\sum_{m=1}^k\frac{b_m^2(n+m-1)}{m q_m}=V_q.
\]
For the distribution in Eq.~\eqref{eq:bandit-sampling-distribution},
$V_q\leq32e\,nk^2$.
\end{restatedlemma}

\begin{proof}
Conditional on $M_t=m$,
$N_{t,i}\sim\mathrm{Binomial}(m,p_{t,i})$. Hence
\[
    \E_t[N_{t,i}^2\mid M_t=m]
    =mp_{t,i}+m(m-1)p_{t,i}^2.
\]
Dividing by $p_{t,i}$ and summing over $i$ gives
\[
    \E_t\!\left[
        \sum_{i=1}^n\frac{N_{t,i}^2}{p_{t,i}}
        \,\middle|\,M_t=m
    \right]
    =mn+m(m-1)=m(n+m-1).
\]
Since $(f_t(S_t)-1)^2\leq1$,
\[
    \E_t\!\left[
        \sum_{i=1}^n p_{t,i}z_{t,i}^2
        \,\middle|\,M_t=m
    \right]
    \leq
    \frac{a_m^2}{q_m^2}m(n+m-1).
\]
Averaging over $M_t$ and using $a_m=b_m/m$ proves
\[
    \E_t\!\left[\sum_{i=1}^n p_{t,i}z_{t,i}^2\right]
    \leq
    \sum_{m=1}^k\frac{a_m^2}{q_m}m(n+m-1)
    =V_q.
\]

For the chosen distribution,
$b_m\leq\sum_{\ell=m}^kq_\ell\leq1$ and
$b_k=q_k=1-\beta$. Therefore,
\begin{align*}
    V_q
    &\leq
    (n+k-1)\left(
        \sum_{m=1}^{k-1}\frac1{m q_m}
        +\frac{1-\beta}{k}
    \right)\\
    &=
    (n+k-1)\left(
        \frac{Z_{k-1}^2}{\beta}
        +\frac{1-\beta}{k}
    \right).
\end{align*}
This also explains the allocation of probability among $m<k$:
by Cauchy--Schwarz, for any positive probabilities with
$\sum_{m<k}q_m=\beta$,
\[
    Z_{k-1}^2
    =
    \left(\sum_{m=1}^{k-1}
        \sqrt{q_m}\,\frac1{\sqrt{m q_m}}\right)^2
    \leq
    \beta\sum_{m=1}^{k-1}\frac1{m q_m}.
\]
Equality holds when $q_m=\beta/(Z_{k-1}\sqrt m)$.

Finally, an integral comparison gives
$Z_{k-1}\leq2\sqrt{k-1}$. Also,
$\log(1-1/k)\leq-1/k-1/(2k^2)$ implies
\[
    \beta
    =\frac1e-\rho^k
    \geq
    \frac1e(1-e^{-1/(2k)})
    \geq\frac1{4ek},
\]
using $1-e^{-x}\geq x/2$ for $0\leq x\leq1$.
It follows that
\[
    V_q
    \leq
    (n+k-1)\left(16e\,k(k-1)+\frac1k\right)
    \leq16e\,k^2(n+k-1)
    \leq32e\,nk^2.
\]
\end{proof}

\subsection{Proof of Theorem~\ref{thm:bandit-entropy}}

\begin{restatedtheorem}[Restatement of Theorem~\ref{thm:bandit-entropy}, Exponential-weights guarantee]
Let $2\leq k<n$, and let $f_1,\ldots,f_T:2^V\to[0,1]$ be normalized,
monotone, submodular functions fixed in advance by an oblivious adversary.
For every $\eta>0$, Algorithm~\ref{alg:bandit} satisfies
\[
    \aReg
    \leq\frac{\log(n/k)}{\eta}+\frac{\eta T V_q}{2}.
\]
In particular, choosing $\eta=\sqrt{2\log(n/k)/(T V_q)}$ gives
\[
    \aReg
    \leq\calO\!\left(k\sqrt{nT\log(n/k)}\right).
\]
\end{restatedtheorem}

\begin{proof}
We use the second-order exponential-weights bound with a
distributional comparator $\bu\in\Delta_n$.
For completeness, define
\[
    D_{\mathrm{KL}}(\bu\|\bv)
    \coloneq
    \sum_{i:u_i>0}u_i\log\frac{u_i}{v_i},
    \qquad
    W_t\coloneq\sum_{i=1}^n p_{t,i}\exp(\eta z_{t,i}).
\]
The update gives
\[
    D_{\mathrm{KL}}(\bu\|\bp_{t+1})
    -D_{\mathrm{KL}}(\bu\|\bp_t)
    =\log W_t-\eta\langle\bz_t,\bu\rangle.
\]
Since $z_{t,i}\leq0$, the inequalities
$e^x\leq1+x+x^2/2$ for $x\leq0$ and
$\log x\leq x-1$ give
\[
    \log W_t
    \leq
    \eta\langle\bz_t,\bp_t\rangle
    +\frac{\eta^2}{2}\sum_{i=1}^n p_{t,i}z_{t,i}^2.
\]
Rearranging, summing over $t$, and using nonnegativity of
KL divergence yields
\[
    \sum_{t=1}^T
    \langle\bz_t,\bu-\bp_t\rangle
    \leq
    \frac{D_{\mathrm{KL}}(\bu\|\bp_1)}{\eta}
    +\frac{\eta}{2}\sum_{t=1}^T\sum_{i=1}^n p_{t,i}z_{t,i}^2.
\]
This is the KL-divergence form of the second-order
exponential-weights analysis; see also
\cite[Theorem~1.5]{hazan2016introduction}.
The inequality holds for every realization of the feedback
and every $\eta>0$.

For a set $S$ of size $k$, $\bp^S$ is uniform on $S$ and
$\bp_1$ is uniform on $V$, so
\[
    D_{\mathrm{KL}}(\bp^S\|\bp_1)
    =
    \sum_{i\in S}\frac1k\log\frac{1/k}{1/n}
    =\log(n/k).
\]
Taking expectations and applying
Lemma~\ref{lem:bandit-weighted-moment} therefore gives, for
every such $S$,
\[
    \E\!\left[
        \sum_{t=1}^T
        \langle\bz_t,\bp^S-\bp_t\rangle
    \right]
    \leq
    \frac{\log(n/k)}{\eta}
    +\frac{\eta T V_q}{2}.
\]
Apply Theorem~\ref{thm:bandit-olo} to obtain the first claim.
With the stated learning rate, the bound becomes
$\sqrt{2T V_q\log(n/k)}$; the estimate
$V_q\leq32e\,nk^2$ from
Lemma~\ref{lem:bandit-weighted-moment} gives the second claim.
\end{proof}

\section{A Projected OGD Variant}
\label{app:bandit-second-moments}
\label{sec:bandit-ogd}

This appendix gives an alternative Euclidean instantiation of the
bandit reduction. We keep $q$, $a_m$, $b_m$, $\rho$, and $\beta$
from the main section, and use the same sampling rule and feedback
vector $z_t$ from Eq.~\eqref{eq:bandit-feedback}. Unlike exponential
weights, the Euclidean analysis needs a uniform bound on inverse
coordinate probabilities. We therefore work on a restricted simplex.

\subsection{Restriction and the smoothed comparator}

Set
\[
 \epsilon_0\coloneq\beta\rho^k,\qquad
 K\coloneq\Delta_n^{\epsilon_0}
   \coloneq\{p\in\Delta_n:p_i\ge\epsilon_0/n\text{ for all }i\in V\}.
\]
For $S\subseteq V$ with $|S|=k$, define
\[
 p^{S,\epsilon_0}\coloneq(1-\epsilon_0)p^S+\epsilon_0\frac{\mathbf1}{n}
 \in K.
\]
We first show that smoothing still permits the target benchmark.

\begin{lemma}[Smoothed linearization inequality]
\label{lem:bandit-smoothed-comparator}
For any $\epsilon\in[0,1)$, let
$p^{S,\epsilon}\coloneq(1-\epsilon)p^S+\epsilon\mathbf1/n$.
For every $p\in\Delta_n$,
\[
 (1-\epsilon)\alpha(q)f_t(S)-H_t(p)
 \le\langle\nabla G_t(p),p^{S,\epsilon}-p\rangle.
\]
For the choices of $q$ and $\epsilon_0$ above,
$(1-\epsilon_0)\alpha(q)\ge1-1/e$.
\end{lemma}

\begin{proof}
For every set $B$, submodular averaging and monotonicity give
\[
 \begin{aligned}
 \sum_i p_i^{S,\epsilon}f_t(B\cup\{i\})
 &\ge(1-\epsilon)\left(\frac1k f_t(S)+\rho f_t(B)\right)
      +\epsilon f_t(B)\\
 &=\frac{1-\epsilon}{k}f_t(S)
      +\left(\rho+\frac\epsilon k\right)f_t(B).
 \end{aligned}
\]
Using the gradient identity and summing with weights $a_m=b_m/m$,
the same cancellation as in Lemma~\ref{lem:bandit-comparator} yields
\[
 \begin{aligned}
 \langle\nabla G_t(p),p^{S,\epsilon}-p\rangle
 \ge{}&(1-\epsilon)\alpha(q)f_t(S)-H_t(p)\\
       &+\frac\epsilon k\sum_{m=1}^k b_mF_{t,m-1}(p).
 \end{aligned}
\]
The last term is nonnegative. For the coefficient, Lemma~\ref{lem:q-choice}
gives $\alpha(q)\ge1-1/e+\epsilon_0$. Since $\alpha(q)\le1$,
\[
 (1-\epsilon_0)\alpha(q)
 \ge\alpha(q)-\epsilon_0\ge1-\frac1e.
\]
Also $0<\epsilon_0<1$, so $K$ is nonempty and contains the uniform
distribution.
\end{proof}

Combining the lemma with $\E_t[f_t(S_t)]=H_t(p_t)$ and
Lemma~\ref{lem:estimator-mean}, for iterates in $K$ we obtain
\begin{equation}
 \left(1-\frac1e\right)f_t(S)-\E_t[f_t(S_t)]
 \le\E_t[\langle z_t,p^{S,\epsilon_0}-p_t\rangle].
 \label{eq:ogd-smoothed-reduction}
\end{equation}
The restriction is independent of $T$: its cost is absorbed by the
approximation coefficient rather than by an additive per-round error.

\subsection{Algorithm and Euclidean second moment}

Initialize $p_1=\mathbf1/n$. On each round, sample $M_t$, the draws,
and the feedback $z_t$ exactly as in Algorithm~\ref{alg:bandit}, but
replace its exponential-weights update by
\begin{equation}
 p_{t+1}\coloneq\Pi_K(p_t+\eta z_t),
 \label{eq:bandit-ogd-update}
\end{equation}
where $\Pi_K$ is Euclidean projection onto $K$. This is projected
gradient ascent on the linear reward $\langle z_t,p\rangle$,
or equivalently OGD on the linear loss $-\langle z_t,p\rangle$.
Using $\widehat g_t=z_t+c_q\mathbf1$ gives the same projection,
since every point of $K$ has coordinate sum one.

\begin{lemma}[Euclidean second moment]
\label{lem:bandit-euclidean-moment}
If $p_t\in K$, then
\[
 \E_t[\|z_t\|_2^2]\le\frac{nV_q}{\epsilon_0}.
\]
\end{lemma}

\begin{proof}
The coordinate floor gives, pointwise,
\[
 \sum_i z_{t,i}^2\le\frac n{\epsilon_0}
                 \sum_i p_{t,i}z_{t,i}^2.
\]
Take conditional expectations and apply
Lemma~\ref{lem:bandit-weighted-moment}.
\end{proof}

\subsection{Regret guarantee}

\begin{theorem}[Projected OGD guarantee]
\label{thm:bandit-ogd}
Under the setting of the main bandit section, the projected OGD
variant above satisfies, for every $\eta>0$,
\[
 \aReg
 \le\frac1{2k\eta}+\frac{\eta nTV_q}{2\epsilon_0}.
\]
In particular, choosing $\eta=\sqrt{\epsilon_0/(knTV_q)}$ gives
\[
 \aReg
 \le\sqrt{\frac{nTV_q}{k\epsilon_0}}
 =\calO\!\left(nk\sqrt T\right).
\]
\end{theorem}

\begin{proof}
For every $u\in K$, nonexpansiveness of Euclidean projection gives
\[
 \begin{aligned}
 \|p_{t+1}-u\|_2^2
 &\le\|p_t+\eta z_t-u\|_2^2\\
 &=\|p_t-u\|_2^2-2\eta\langle z_t,u-p_t\rangle
       +\eta^2\|z_t\|_2^2.
 \end{aligned}
\]
Rearranging, summing, and dropping the final squared distance yields
\[
 \sum_{t=1}^T\langle z_t,u-p_t\rangle
 \le\frac{\|u-p_1\|_2^2}{2\eta}
       +\frac\eta2\sum_{t=1}^T\|z_t\|_2^2.
\]
Taking expectations and applying
Lemma~\ref{lem:bandit-euclidean-moment},
\[
 \E\!\left[\sum_{t=1}^T\langle z_t,u-p_t\rangle\right]
 \le\frac{\|u-p_1\|_2^2}{2\eta}
       +\frac{\eta nTV_q}{2\epsilon_0}.
\]
For $u=p^{S,\epsilon_0}$ and $p_1=\mathbf1/n$,
\[
 \|u-p_1\|_2^2
 =(1-\epsilon_0)^2\left(\frac1k-\frac1n\right)\le\frac1k.
\]
Taking expectations and summing
Eq.~\eqref{eq:ogd-smoothed-reduction} therefore proves the claimed
regret bound for every fixed $S$ of size $k$. Maximizing over $S$
and optimizing in $\eta$ gives the explicit square-root guarantee.

The main proof established $V_q=\calO(nk^2)$ and
$\beta^{-1}=\calO(k)$. Moreover,
$\log(1-x)\ge-x/(1-x)$ for $0\le x<1$ implies
\[
 \rho^k\ge\exp\!\left(-\frac{k}{k-1}\right)\ge e^{-2}.
\]
Thus $\epsilon_0^{-1}=(\beta\rho^k)^{-1}=\calO(k)$,
which gives the stated dependence on $n,k,T$.
\end{proof}

\paragraph{Computational complexity.}
Write $d\coloneq\epsilon_0/n$. The projection has the form
\[
 [\Pi_K(v)]_i=\max\{d,v_i-\tau\},
 \qquad \sum_i\max\{d,v_i-\tau\}=1,
\]
where the threshold $\tau$ can be found by sorting the coordinates.
This takes $O(n\log n)$ time. Including the sampling and estimator
construction, the OGD variant takes $O(n\log n)$ time per round
because $k\le n$, and $O(n+k)$ storage, in the same arithmetic model
as the main algorithm.

\section{The Exponential Weights Algorithm}
\label{app:exponential-weights}

Exponential weights maintains a distribution $\bp_t\in\Delta_n$
over $n\geq2$ coordinates, assigning greater probability to coordinates
with larger cumulative feedback. Starting from
$\bp_1=\bone/n$, it uses learning rate $\eta>0$ and updates
\begin{equation}
    p_{t+1,i}
    =\frac{p_{t,i}\exp(\eta v_{t,i})}
    {\sum_{j=1}^n p_{t,j}\exp(\eta v_{t,j})}.
    \label{eq:app-ew-update}
\end{equation}
The feedback $\bv_t$ depends on the observation model.
For a fixed distribution $\bu\in\Delta_n$, write
$D_{\mathrm{KL}}(\bu\|\bp)
:=\sum_{i:u_i>0}u_i\log(u_i/p_i)$.
All logarithms are natural.

\subsection{Full-information feedback}
\label{app:ew-full-information}

After choosing $\bp_t$, the learner observes the entire reward
vector $\bg_t\in[0,B]^n$, where $B>0$, and sets
$\bv_t=\bg_t$. Its linear reward is
$\langle\bg_t,\bp_t\rangle$.
This is the Hedge algorithm; see~\citet{hazan2016introduction}(Section~1.3.3).
We state the standard regret bound in its Hoeffding-lemma form.

\begin{theorem}[Full-information regret]
\label{thm:app-ew-full-information}
For any sequence $\bg_t\in[0,B]^n$ and any $\eta>0$,
exponential weights satisfies, for every fixed distribution
$\bu\in\Delta_n$,
\begin{equation}
    \sum_{t=1}^T
    \langle\bg_t,\bu-\bp_t\rangle
    \leq\frac{D_{\mathrm{KL}}(\bu\|\bp_1)}{\eta}
        +\frac{\eta B^2T}{8}.
    \label{eq:app-ew-full-information-bound}
\end{equation}
Since $D_{\mathrm{KL}}(\bu\|\bp_1)\leq\log n$,
choosing $\eta=\sqrt{8\log n/(B^2T)}$ bounds regret against the best
fixed action by $B\sqrt{T\log n/2}$. This is also the bound against
the best fixed distribution, since rewards are linear.
\end{theorem}

Our full-information reduction uses
$\bg_t=\nabla G_t(\bp_t)\in[0,k]^n$;
setting $B=k$ recovers the guarantee in the main text.

\subsection{Bandit feedback}
\label{app:ew-bandit}

With bandit feedback, the learner observes only the reward of its sampled
action and constructs an importance-weighted estimate. This is the
principle underlying Exp3, introduced by~\cite{auer2002nonstochastic}.
We use the loss-based formulation described by~\citet{hazan2016introduction}(Section~6.2.1), expressed as nonpositive rewards.
For example, when $I_t\sim\bp_t$ and only $g_{t,I_t} \in [0, B]$ is observed,
set $\bv_t=\bz_t$, where
\[
    z_{t,i}:=\frac{(g_{t,I_t}-B)\mathbf{1}\{I_t=i\}}{p_{t,i}}.
\]
Then $\mathbb{E}_t[\bz_t]=\bg_t-B\bone$.
Here $\mathbb{E}_t$ conditions on the history and the current reward
vector, determined before sampling. The common shift leaves linear
comparisons unchanged because
$\langle\bone,\bu-\bp_t\rangle=0$.

The following second-moment form of the exponential-weights guarantee
also accommodates the set-reward estimator used in our reduction.
It follows from the standard Hedge/Exp3 analysis in~\citet{hazan2016introduction}(Sections~1.3.3 and~6.2.1), retaining
$D_{\mathrm{KL}}(\bu\|\bp_1)$ in the bound
and using $e^x\leq1+x+x^2/2$ for $x\leq0$.
Thus, no uniform bound on the magnitude of the nonpositive estimates
is required.

\begin{theorem}[Bandit regret]
\label{thm:app-ew-bandit}
Suppose the update uses finite feedback $\bz_t\leq0$ with
\[
    \mathbb{E}_t[\bz_t]
      =\bg_t-c_t\bone,
    \qquad
    \mathbb{E}_t\!\left[\sum_i p_{t,i}z_{t,i}^2\right]\leq V,
\]
where $c_t$ is determined before the current sample and $V>0$.
For every fixed distribution $\bu\in\Delta_n$ and every $\eta>0$,
\begin{equation}
    \mathbb{E}\!\left[\sum_{t=1}^T
      \langle\bg_t,\bu-\bp_t\rangle
    \right]
    \leq\frac{D_{\mathrm{KL}}(\bu\|\bp_1)}{\eta}
         +\frac{\eta TV}{2}.
    \label{eq:app-ew-bandit-bound}
\end{equation}
For any collection of fixed distributions satisfying
$D_{\mathrm{KL}}(\bu\|\bp_1)\leq D$, where $D>0$, choosing
$\eta=\sqrt{2D/(TV)}$ gives expected regret at most $\sqrt{2DTV}$
against each distribution in the collection.
\end{theorem}

For the single-arm estimator above,
$\mathbb{E}_t[\sum_i p_{t,i}z_{t,i}^2]
=\sum_i(g_{t,i}-B)^2\leq nB^2$.
Taking $D=\log n$ therefore bounds regret against the best fixed action:
\[
    \max_{i\in[n]}\sum_{t=1}^T g_{t,i}
    -\mathbb{E}\!\left[\sum_{t=1}^T g_{t,I_t}\right]
    \leq B\sqrt{2nT\log n}
\]
for an oblivious reward sequence.

In our bandit reduction, the estimator satisfies the theorem with
$\bg_t=\nabla G_t(\bp_t)$,
$c_t=\sum_m b_m$, and $V=V_q$, as established in the main text.
For the uniform distribution $\bp^S$ on a fixed $k$-element set $S$,
$D_{\mathrm{KL}}(\bp^S\|\bp_1)=\log(n/k)$,
which yields the bound $\log(n/k)/\eta+\eta TV_q/2$ used there
to control $(1-1/e)$-regret against the best fixed set.

\section{Experiments}
\label{app:experiments}

Our code and data are available on \href{https://github.com/jasonNikolaou/bandit-osm}{GitHub}.

\subsection{Datasets and Problem Instances}
\label{subsec:datasets}

We evaluate the algorithms on three monotone submodular maximization
problems: a weighted influence-maximization instance on
\emph{Zachary's Karate Club} graph~\citep{zachary1977information},
\Influence; a synthetic team-formation instance, \TeamFormation; and a
synthetic weighted-coverage instance, \Coverage.

At every round $t$, the learner selects a set $S_t\subseteq[n]$ satisfying
$|S_t|\leq k$ and observes only the scalar reward $f_t(S_t)$. The learner
does not observe $f_t$, its marginal gains, or the rewards of sets that
were not selected. All objectives are monotone and submodular. Rewards
are normalized to lie in $[0,1]$ using a common valid upper bound for each
problem instance.

For \TeamFormation and \Coverage, we generate ten base task types and
sample the task type at each round independently and uniformly. For
\Influence, we independently sample a new live-edge graph at every
round. Table~\ref{tab:datasets} summarizes the instance parameters.

\begin{table}[!t]
\centering
\begin{tabular}{lrrrr}
\toprule
\textbf{Dataset} & $n$ & $k$ & \textbf{Task types} & $T$ \\
\midrule
\Influence     & 34  & 2 & Fresh live-edge graph & $10{,}000$ \\
\TeamFormation & 100 & 3 & 10                    & $100{,}000$ \\
\Coverage      & 100 & 3 & 10                    & $100{,}000$ \\
\bottomrule
\end{tabular}
\caption{Parameters of the experimental problem instances.}
\label{tab:datasets}
\end{table}

\spara{Influence.}
The \Influence instance is defined on \emph{Zachary's Karate Club}
graph, which contains $n=34$ vertices and $78$ undirected edges. At each
round, every edge is independently retained with probability $p=0.05$,
producing a live-edge realization of the graph.

The reward of a seed set $S$ is the total weight of the vertices reachable
from at least one seed in $S$. Equivalently, it is the total weight of the
connected components intersected by $S$. This is a weighted coverage
function and is therefore monotone and submodular~\citep{kempe2003maximizing}.

Vertices $0$ and $33$ receive weight $0.5$, while every other vertex
receives weight $0.01$. We use the cardinality constraint $k=2$ and
normalize rewards by the total vertex weight, which equals $1.32$.

\spara{Team Formation.}
The \TeamFormation instance models roster selection over a pool of
$n=100$ candidates. The learner selects a team of at most $k=3$
candidates. For task type $\tau$, the reward is the monotone submodular
quadratic function
\begin{align*}
    f_\tau(\by)
    =
    \bw_\tau^\intercal\by
    +
    \frac{1}{2}\by^\intercal\bH_\tau\by,
\end{align*}
where $\by\in\{0,1\}^n$ indicates the selected team,
$\bw_\tau\in\mathbb{R}_{\geq0}^n$ contains the individual candidate
values, and $\bH_\tau$ models redundancy between candidates with similar
skills.

Each candidate $i$ is assigned a fixed embedding
$\bs_i\in\mathbb{R}^8$, sampled from $\mathcal{N}(0,I_8)$. We sample a
common unit requirement vector $\bar{\br}$ and define the requirement
vector of task type $\tau$ as
\begin{align*}
    \br_\tau
    =
    \frac{\bar{\br}+0.05\,\boldsymbol{\xi}_\tau}
    {\|\bar{\br}+0.05\,\boldsymbol{\xi}_\tau\|_2},
\end{align*}
where $\boldsymbol{\xi}_\tau$ is an independently sampled unit vector.
The ten task types therefore have related, but nonidentical, skill
requirements.

The raw score of candidate $i$ on task $\tau$ uses its normalized
embedding:
\begin{align*}
    40\left\langle
        \frac{\bs_i}{\|\bs_i\|_2},\br_\tau
    \right\rangle+\epsilon_{\tau,i},
    \qquad
    \epsilon_{\tau,i}\sim\mathcal{N}(0,1).
\end{align*}
The scores are shifted and rescaled to $[0,100]$. We designate $75\%$
of the candidates as permanently low-value candidates. Among the
remaining candidates, only the ten highest-scoring candidates for each
task retain their similarity-based scores. Every other candidate receives
a small value sampled from $[0.05,0.5]$.

To model overlap between candidates, we construct a
$15$-nearest-neighbor graph using cosine similarity between their
embeddings. For every edge $(i,j)$, we set
\begin{align*}
    (\bH_\tau)_{ij}
    =
    (\bH_\tau)_{ji}
    =
    -\lambda_\tau
    \max\{0,\cos(\bs_i,\bs_j)\}
    \frac{\min\{w_{\tau,i},w_{\tau,j}\}}{k-1},
\end{align*}
where $\lambda_\tau$ is sampled uniformly from $[0.05,0.15]$. All
remaining entries are zero. The nonpositive pairwise terms imply
submodularity. For any feasible addition to a set of fewer than $k$
candidates, the total reduction in the marginal value of candidate $i$
is at most $\lambda_\tau w_{\tau,i}<w_{\tau,i}$, so the objective is
monotone on the feasible region. For the fixed evaluation and tuning
instances, we also verified
$w_{\tau,i}+\sum_{j\ne i}(\bH_\tau)_{ij}>0$ for every task and candidate;
hence their objectives are monotone on all subsets.

We sample a length-$T=10^5$ stream independently and uniformly from the
ten task types. Rewards are normalized by the largest, across task types,
sum of the $k$ largest individual candidate values. This is a valid upper
bound because the quadratic term is nonpositive.

\spara{Coverage.}
The \Coverage instance is a weighted coverage problem with $n=100$
selectable elements, cardinality constraint $k=3$, and a universe of
$250$ features. The universe consists of ten topics with $20$ features
each and $50$ low-value background features. Each element $i$ covers a
fixed subset $C_i$ of the feature universe. For task type $\tau$, the
reward is
\begin{align*}
    f_\tau(S)
    =
    \sum_{u\in\bigcup_{i\in S}C_i}a_{\tau,u},
\end{align*}
where $a_{\tau,u}\geq0$ is the task-dependent weight of feature $u$.
This is a monotone submodular weighted coverage function.

We generate ten related task types. Three topics are shared across all
tasks, with feature weights sampled from $[0.8,1.2]$. Each task also
emphasizes one of the remaining seven topics, whose feature weights are
sampled from $[0.15,0.30]$. Features in all other topics receive weights
in $[0.01,0.03]$, while background features receive weights in
$[0.001,0.005]$.

We designate $75\%$ of the elements as low-value elements, each covering
only two background features. Of the remaining elements, ten are topic
specialists. Each specialist covers $16$ features from one topic and two
background features. Ten shared generalists each cover eight features
from every shared topic and two background features. The final five
generalists cover three features from each of three randomly selected
topics and two background features.

We sample a length-$T=10^5$ stream independently and uniformly from the
ten task types. Rewards are normalized by the maximum total feature
weight of any task type. This is a valid upper bound because no selected
set can cover more than the complete feature universe.

\subsection{Algorithms}
\label{app:algorithms}

\spara{Sub-EW.}
We evaluate our bandit algorithm, denoted by {\alg}. At each round,
{\alg} samples a draw count from the distribution prescribed by
Algorithm~\ref{alg:bandit}, draws that many elements independently with
replacement from its current exponential-weights distribution, and
plays their union. It updates its distribution using only the normalized
reward of the selected set and the importance-weighted estimator
developed in the paper.

\spara{Baselines.}
We compare {\alg} with six baselines:
\begin{itemize}
    \item \Random selects a uniformly random subset of exactly $k$
    distinct elements at every round.

    \item \MSEThree implements Algorithm~1 of \citet{pasteris2023sum}.
    Like {\alg}, it maintains one item distribution and samples
    with replacement, but always makes $k$ draws.
    It estimates the gradient of the fixed-draw expected reward,
    and its guarantee relies on pseudo-concavity.

    \item \OpaqueGreedy implements the opaque-feedback greedy algorithm
    of \citet{streeter2008online}. It maintains one weighted-majority
    learner per greedy stage and uses prefix-plus-element exploration.

    \item \BanditIG implements Algorithms~3--4 and Example~1 of
    \citet{niazadeh2023online}. Each greedy stage has a Hedge learner
    that instantiates the required Blackwell-approachability subroutine.
    After its $r$th feedback update, that learner uses weights
    $w_{r+1,i}\propto
    \exp(-\eta_r\sum_{s=1}^{r}\widehat\ell_{s,i})$.

    \item \BanditMLSM specializes Algorithms~2 and~4 of~\citet{wan2023bandit} to one partition group of capacity $k$. Its product
    of $k$ simplexes includes a dummy option in each factor, so a
    sampled set contains at most $k$ elements. One exploration round
    per block provides a scalar-feedback gradient estimate for its
    log-barrier follow-the-regularized-leader update.

    \item \PoissonBE specializes Algorithm~2 of~\citet{wan2026bandit}
    to the uniform rank-$k$ matroid. It uses a Poisson base walk and
    balanced exchanges, estimates an insertion loss from one
    feasible-set reward, and applies an entropy update followed by
    projection onto a shrunken base polytope.
\end{itemize}
Every algorithm observes only the normalized reward of its selected set.

\spara{Hyperparameters.}
For {\alg}, \MSEThree, \OpaqueGreedy, and \BanditIG, we multiply
the prescribed learning rate by
$c_\eta\in\{0.25,1,4,16,64\}$. 
For each dataset and algorithm,
we select the candidate with the highest mean normalized reward over three independent repetitions on a separately generated tuning instance. The tuning horizon is $\min\{T,10^4\}$. The multiplier $c_\eta=1$ is evaluated first and wins ties.

\OpaqueGreedy uses exploration probability
$\gamma=\min\{1,k(n\log n/T)^{1/3}\}$, and \BanditIG uses
$q=\min\{1,(n^2\log n/T)^{1/3}\}$. For \BanditIG, $T$ in this
formula and in its Hedge schedule is the evaluation horizon, even
when candidates are scored on shorter tuning streams. The selected
multipliers are reported in Table~\ref{tab:learning_rates}.

BanditMLSM and Poisson-BE use separate practical tuning grids.
For BanditMLSM, we tune the block length
$L\in\{8,32,128,512\}$ and the learning rate
$\eta=c_\eta d^{-4}T^{-2/3}$, where $d=nk$. The unscaled rate
comes from Wan et al.'s Algorithm~1 and serves only as a practical
tuning reference for our product-simplex implementation. The
multiplier grid is $\{10^4,10^5,10^6,10^7\}$ for \Influence and
$\{10^6,10^7,10^8\}$ for \TeamFormation and \Coverage.
For Poisson-BE, we tune exploration probability
$\gamma\in\{0.05,0.2,0.5\}$ and a multiplier
$c_\eta\in\{0.25,1,4,16,64,256\}$ of
$\sqrt{k^2\gamma\log(en)/(2n(\mu^2+\mu)T)}$, where
$\mu=k\log T$. Its starting time $\varepsilon=1/T$,
shrinkage $\delta=\min\{1/2,1/(\mu T)\}$, and Poisson event cap
$\lceil 2e(\mu+\log T)\rceil$ follow the published formulas.
Both methods use the evaluation horizon to compute their parameter
values and three independent shorter tuning streams for candidate
selection.

For the product-simplex wrapper, the proof of Lemma~5.1 in
Wan et al. uses $\eta=d^{-7/3}T^{-1/3}$ and
$L=d^{-5/3}T^{1/3}$. The latter is below one round at our
dimensions. Poisson-BE's learning-regime condition
$n(\mu^2+\mu)\log(en)<T$ also does not hold. The tuned
configurations retain the published sampling and update rules;
they are practical instantiations rather than finite-horizon
applications of those theorems.
Table~\ref{tab:wan_parameters} reports their selected parameters.

\begin{table}[!t]
\centering
\begin{tabular}{lrrrr}
\toprule
\textbf{Dataset}
& {\alg} & \MSEThree & \OpaqueGreedy & \BanditIG \\
\midrule
\Influence     & 4 & 4 & 16 & 64 \\
\TeamFormation & 4 & 4 & 64 & 64 \\
\Coverage      & 4 & 4 & 64 & 64 \\
\bottomrule
\end{tabular}
\caption{Selected multipliers applied to the base learning rates of
{\alg}, \MSEThree, \OpaqueGreedy, and \BanditIG.}
\label{tab:learning_rates}
\end{table}

\begin{table}[!t]
\centering
\begin{tabular}{lrrrr}
\toprule
\textbf{Dataset}
& \multicolumn{2}{c}{BanditMLSM}
& \multicolumn{2}{c}{Poisson-BE} \\
& $L$ & $c_\eta$ & $\gamma$ & $c_\eta$ \\
\midrule
\Influence     & 128 & $10^7$ & 0.2 & 16 \\
\TeamFormation & 512 & $10^6$ & 0.2 & 256 \\
\Coverage      & 512 & $10^8$ & 0.2 & 256 \\
\bottomrule
\end{tabular}
\caption{Practical parameters selected for BanditMLSM and Poisson-BE
on independent tuning streams.}
\label{tab:wan_parameters}
\end{table}

\subsection{Evaluation Protocol}
\label{app:evaluation}

\spara{Average approximate regret.}
For each evaluation stream and every round $t$, let
\begin{align*}
    \mathrm{OPT}_t
    =
    \max_{S:\,|S|\leq k}
    \sum_{s=1}^{t} f_s(S).
\end{align*}
The maximization selects one fixed set for the entire prefix of length
$t$; the maximizing set may differ between prefixes. We report
\begin{align*}
    \mathrm{AAR}(t)
    =
    \frac{1}{t}
    \left[
        (1-1/e)\mathrm{OPT}_t
        -
        \sum_{s=1}^{t}f_s(S_s)
    \right].
\end{align*}
Lower values are better. A negative value means that the algorithm's
cumulative reward exceeds $(1-1/e)\mathrm{OPT}_t$.

We compute $\mathrm{OPT}_t$ at every timestep by exhaustive enumeration
with cumulative candidate scores. Monotonicity permits enumeration of
sets of size exactly $k$: there are $\binom{34}{2}=561$ candidates for
\Influence and $\binom{100}{3}=161{,}700$ for each of \TeamFormation and
\Coverage. For the latter two datasets, we precompute each candidate's
reward for the ten task types. A common maximizer of all task types is
used to obtain the prefix trajectory only after verifying that property
for the loaded instance; otherwise, we update all candidate scores at
each prefix. The metric is computed at every timestep, even when a
figure displays a subsampled curve.

\spara{Repetitions.}
We perform ten independent evaluation repetitions for every dataset.
Within a repetition, all algorithms are evaluated on the same task
stream, while each algorithm uses its own random seed. Across
repetitions, both the task stream and the algorithmic random seeds
change. The underlying graph, candidate embeddings, task types, and
coverage sets remain fixed.

The reported curve is the pointwise mean of $\mathrm{AAR}(t)$ over the
ten evaluation repetitions. The shaded region extends one sample
standard deviation above and below the mean.

\spara{Reproducibility.}
The implementation uses Python~3.10.10, NumPy~2.2.6, SciPy~1.15.3,
NetworkX~3.4.2, and Matplotlib~3.10.9. The repository contains the
source code, fixed problem instances, configuration files, tests, and
plotting scripts.

\spara{Setup.}
All experiments were run on an Apple M2 MacBook with 16 GB of RAM.

\end{document}